\documentclass[10pt,journal,compsoc]{IEEEtran}
\usepackage[colorlinks,anchorcolor=blue,citecolor=blue,linkcolor=blue]{hyperref}
\usepackage{cite}
\usepackage{colortbl}
\usepackage{amsmath}
\usepackage{amssymb}
\usepackage{enumitem}
\usepackage{graphicx}
\usepackage{ragged2e}  
\usepackage[caption=false,font=normalsize,labelfont=sf,textfont=sf,subrefformat=parens,labelformat=parens]{subfig}
\usepackage{makecell}

\usepackage{amsmath,amsfonts,amssymb}
\usepackage{array}
\usepackage[caption=false,font=normalsize,labelfont=sf,textfont=sf]{subfig}
\usepackage{textcomp}
\usepackage{stfloats}
\usepackage{url}
\usepackage{verbatim}
\def\BibTeX{{\rm B\kern-.05em{\sc i\kern-.025em b}\kern-.08em
    T\kern-.1667em\lower.7ex\hbox{E}\kern-.125emX}}
\usepackage{balance}
\usepackage{booktabs}
\usepackage{graphicx}
\usepackage{multirow}
\usepackage{ntheorem}
\usepackage{bbding}
\usepackage[numbers]{natbib}
\usepackage{algorithm}
\usepackage{algorithmic}

\newcommand{\CASE}[1]{ \STATE  \textbf{case} #1\textbf{:} \begin{ALC@g}}
\newcommand{\ENDCASE}{\end{ALC@g}}

\newcommand{\DEFAULT}{\STATE \textbf{default:} \begin{ALC@g}}
\newcommand{\ENDDEFAULT}{\end{ALC@g}}
\newcommand{\DEFAULTLINE}[1]{\STATE \textbf{default:} }

\usepackage{bm}
\usepackage{pifont}
\usepackage[table]{xcolor}
\usepackage{enumitem}
\usepackage{relsize}

\theorembodyfont{\rmfamily}

\newtheorem{definition}{Definition}
\newtheorem{theorem}{Theorem}

\newtheorem{prop}{Proposition}

\newcolumntype{L}{>{\centering\arraybackslash}m{.5cm}}

\newcolumntype{S}{>{\centering\arraybackslash}m{.65cm}}

\DeclareMathOperator*{\argmin}{\arg\!\min}

\begin{document}

\title{Locality-aware Private Class Identification for Domain Adaptation with Extreme Label Shift}

\author{Chuan-Xian~Ren,~Cheng-Jun~Guo,~and~Hong~Yan~\IEEEmembership{Fellow,~IEEE}
\IEEEcompsocitemizethanks{\IEEEcompsocthanksitem C.X. Ren and C.J. Guo are with the School of Mathematics, Sun Yat-Sen University, Guangzhou 510275, China. Hong Yan is with the Department of Electrical Engineering, City University of Hong Kong, Hong Kong, China. C.X. Ren is the corresponding author (email: rchuanx@mail.sysu.edu.cn).\protect\\
\IEEEcompsocthanksitem This work is supported in part by National Key R\&D Program of China (2024YFA1011900), in part by National Natural Science Foundation of China (Grant No. 62376291), in part by Guangdong Basic and Applied Basic Research Foundation (2023B1515020004), in part by Science and Technology Program of Guangzhou (2024A04J6413), in part by the Fundamental Research Funds for the Central Universities, Sun Yat-sen University (24xkjc013), and in part by the Hong Kong Innovation and Technology Commission (ITC) (InnoHK Project CIMDA) and the Institute of Digital Medicine of City University of Hong Kong (Project 9229503).}}



\markboth{IEEE TRANSACTIONS ON PATTERN ANALYSIS AND MACHINE INTELLIGENCE}%
{Shell \MakeLowercase{\textit{\etal}}: Bare Demo of IEEEtran.cls for Computer Society Journals}
\IEEEtitleabstractindextext{
\begin{abstract}
\justifying
Domain adaptation aims to transfer knowledge from a labeled source domain to an unlabeled target domain with different distributions. In real-world scenarios, the label spaces of the two domains often have an inclusion relationship, where some classes exist only in one domain but not the other. These non-overlapping classes are referred to as private classes. Identifying private class samples and mitigating their adverse effects is critical in the literature. Existing methods rely on the assumption that shifts in private classes are large enough to be considered outliers. However, the variance within a single shared class can be significantly larger than the difference between a private class and another shared class, challenging this assumption. Consequently, private classes substantially increase the difficulty of cross-domain classification. To address these issues, based on local transportation and metric properties of optimal transport (OT), a locality-aware private class identification approach is proposed in the form of a score function on transport mass. The effectiveness of the proposed approach is theoretically proven, highlighting the score function's strong ability to distinguish between shared and private class samples. Building on this, we introduce a reliable OT-based method (ReOT) for domain adaptation under severe label shift. ReOT minimizes classification risk while learning the separated cluster structure between the identified shared classes and private classes, effectively avoiding mismatch between shared-private sample pairs, thus ensuring that important knowledge is reliably transported intra-class to mitigate class-conditional discrepancy. Furthermore, a generalization upper bound of the target risk is provided for extreme label shift scenarios, which can be minimized by ReOT. Extensive experiments on benchmarks validate the effectiveness of ReOT.
\end{abstract}

}

\maketitle


\IEEEdisplaynontitleabstractindextext
\IEEEpeerreviewmaketitle

\IEEEraisesectionheading{\section{Introduction}\label{sec:introduction}}

\IEEEPARstart{I}{t} is well-known that a model with strong generalization performance requires large-scale datasets with sufficient label annotations. However, data collected from real world are often unlabeled, and labeling large-scale datasets typically involves excessive costs. In particular, the collected datasets may have different data distributions, resulting in a severe degradation of the model performance. To solve these issues, unsupervised domain adaptation (UDA)~\citep{liu2023cot,kim2024LCMSM,wang2024CMA} has recieved wide attention to transfer knowledge from a labeled domain (i.e., source domain) to an unlabeled domain with different data distribution (i.e., target domain), notably saving time and labor consuming to collect labels for target domain. However, most existing UDA works~\cite{long2016UDA,ganin2016DANN,long2018CDAN} follow a closed set assumption, i.e., the two domains share identical label space. This severely limits their applications to real-world scenarios since the label spaces of source and target domains are usually heterogeneous or extremely shifted. Specifically, the heterogeneous scenarios can be roughly divided into two categories: 1) source label space is a subset of the target label space; 2) source label space includes the target label space. To address these two realistic challenges, open set domain adaptation (OSDA)~\citep{saito2018OSBP,bucci2020ROS,li2023ANNA} and partial domain adaptation (PDA)~\citep{cao2018PADA, luo2022MUL,guo2025IS2C} are proposed, respectively. These two scenarios can be viewed as cases of extreme label shift DA.

Generally, OSDA assumes the private classes exist in the target domain, and it requires the model to recognize them. PDA shows a similar scenario while the private classes exist in the source domain. Therefore, both OSDA and PDA face two primary challenges: 1) Domain gap on shared classes: The data distribution for shared classes differs between the source and target domains, reducing classification accuracy on these classes. 2) Category gap: The absence of private classes in one domain (either source or target) leads to a significant drop in model performance in the target domain, even if the model performs well in the source domain. In these two challenges, the category gap is the core issue. It is not only inherently challenging to be addressed but also complicates the resolution of domain discrepancies using well-studied domain alignment techniques, as the absence of private classes can lead to misalignment between private and shared classes. Therefore, it is crucial to mitigate the adverse effects of the category gap.

Most current works~\citep{liu2019STA,sahoo2023SLM,fang2023PLSC} addressed these challenges by the following two steps: 1) Identifying and then separating the private class samples to eliminate label shift; 2) Implementing domain alignment approach to alleviate domain discrepancy. Following this strategy, these works can recognize some target/source private samples and achieve a degree of domain alignment, thereby obtaining good model performance. While achieving great success, it is worth noting that mainstream methods always focus on only one scenario, ignoring the connection between OSDA and PDA. In fact, OSDA and PDA are symmetric in some sense, as the biggest difference is in which domain the private classes exist. Although OSDA may be more challenging since the private class data are unlabeled while in PDA the private class data are labeled, we would emphasize that a proper method should be able to address both scenarios. 

\begin{figure*}[t] \centering
    \subfloat[Open Set Domain Adaptation (OSDA)]{
        \includegraphics[width=0.47\textwidth]{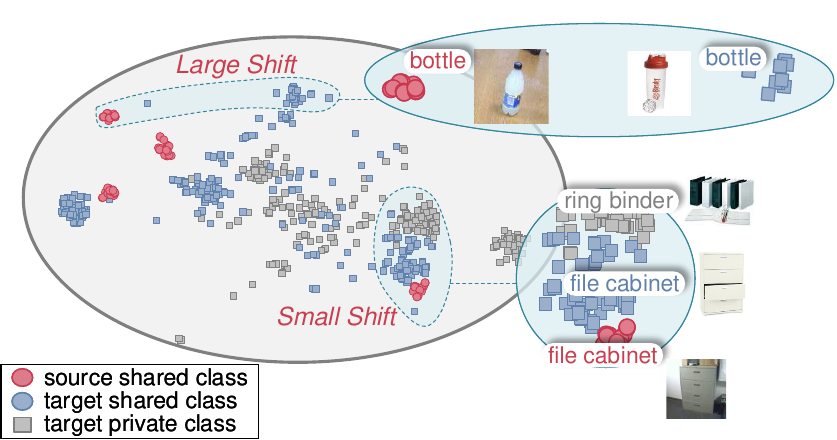}
        \label{fig:motivation_OSDA}}\hfil
    \subfloat[Partial Domain Adaptation (PDA)]{
     \includegraphics[width=0.47\textwidth]{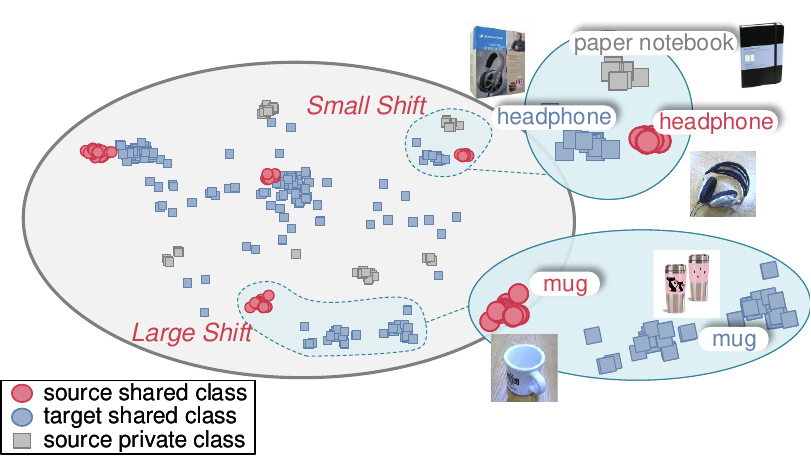}
        \label{fig:motivation_PDA}}
    \caption{Two typical extreme label shifts in unsupervised domain adaptation, i.e., (a) OSDA; (b) PDA. To achieve successful private class identification, existing methods often assume that the semantic shift of the private class is consistently larger than the domain shift of shared classes. However, the variance for a specific shared class can be larger than the gap between a private class and another shared class. Better view in color.}
\label{fig:motivation}.
\end{figure*}

Meanwhile, it is important to recognize that even when considering only one scenario, previous methods have significant limitations, as they often rely on unrealistic strong assumptions to identify private class samples. Specifically, private class identification in previous works mainly includes importance weighting-based approaches~\cite{cao2018PADA,gu2021AR} and threshold-based approaches~\cite{bucci2020ROS,liu2023MRJT,fang2023PLSC}. Importance weighting-based approaches are widely used in PDA. However, they involve assigning very small weights for the private class samples, while learning high-quality weights generally relies on the strong assumption that domain discrepancy is negligible. Threshold-based methods use a metric to determine the likelihood that a sample belongs to a private class. If the metric value exceeds a certain threshold, the sample is classified as private. To ensure effectiveness, these approaches often assume that the gap within a shared class is consistently smaller than the gap between private and shared classes. While this assumption may be less strict than negligible domain discrepancy, it remains difficult to satisfy in practice. As presented in Figure~\ref{fig:motivation}, in real-world scenarios, the gap within a single shared class can significantly exceed the gap between a private class and another shared class. This contradicts the foundational assumptions of many existing methods. Therefore, it is necessary to explore a reliable method, which ensures effectiveness under realistic conditions, for private class identification. 

In the following, we explore the reliability of the private class identification approach for extreme label shift scenarios. We observe that although the shift of shared classes is not always smaller than that of private classes across the entire data manifold, it generally holds true in local regions. Specifically, in a local neighborhood of a shared class cluster, the semantic shift of private class samples is always larger. This observation can be formulated as a local spatial structure. Motivated by this insight, we develop theories to characterize local spatial structures using masked OT~\citep{luo2023MOT}, which offers significant advantages in local correlation characterization and geometric interpretability. Then, we propose a novel locality-aware private class identification method, and theoretically ensure the effectiveness of this method in distinguishing between shared and private class samples. Leveraging this approach, we derive a reliable OT-based model (ReOT) for reliable transfer in practical applications. Furthermore, we provide a generalization upper bound for the target risk in DA scenarios with heterogeneous label spaces, and show that ReOT significantly reduces this bound, indicating its efficiency in addressing extreme label shift DA. Our contributions can be summarized as follows.
\begin{itemize}
    \item A locality-aware private class identification method is proposed using a transport mass-based score function. It effectively captures the complex manifold structure of data by MOT's local transport property and distance-related inequality. MOT's attributes enable our method to mitigate cross-domain class-conditional discrepancy, thus, it is employed for practical knowledge transfer.
    \item The effectiveness of the proposed identification method is theoretically proved, highlighting the score function’s strong ability to distinguish between shared and private class samples. This further implies the reliability of the proposed private class identification approach, i.e., the ability to effectively identify private class samples under realistic conditions.
    \item A theory-guided ReOT method is proposed utilizing locality-aware identification. ReOT minimizes classification risk while learning separated cluster structure with MOT, avoiding mismatch inter-class, and ensuring reliable transfer intra-class. Further, we give a generalization upper bound for extreme label shift scenarios, showing that ReOT effectively minimizes this bound.
    \item Extensive experiments are conducted on several benchmark image datasets, in both OSDA and PDA settings, to validate the empirical effectiveness of ReOT. The results show that ReOT generally outperforms the state-of-the-art methods, demonstrating the stable and superior performance of the learned model.
    
\end{itemize}

The rest of this paper is organized as follows. In Section \ref{sec2}, we provide a brief review of open-set domain adaptation, partial domain adaptation, and the OT-based domain adaptation methods. In Section \ref{sec3}, we present the locality-aware private class identification approach and the ReOT method, following which a generalization error analysis is also provided to show the efficiency of ReOT in addressing extreme label shift DA. Extensive experiments and analysis under standard OSDA/PDA settings are presented in Section \ref{sec4}. Finally, Section \ref{sec5} concludes this paper.



\section{Related Work}\label{sec2}

In this section, several related works on OSDA, PDA and OT-based DA methods are briefly reviewed.

\subsection{Open Set Domain Adaptation} 
OSDA addresses a realistic challenge where the target domain may include private classes. It is difficult for a source pre-trained model to recognize target samples correctly due to the lack of private classes in the source domain.

Saito et al.~\citep{saito2018OSBP} raise the OSDA setting for practical application, where private class samples appear only in the target domain. Recent works generally follow the pipeline to identify private class samples and then align cross-domain shared distributions. Liu et al.~\citep{liu2019STA} propose separate to align (STA), which trains a binary classifier to separate shared and private samples, and implement domain alignment by adversarial learning. Fang et al.~\cite{fang2020DAOD} provide a theoretical bound for open set domain
adaptation, and propose a theory-guided method called distribution alignment with open difference (DAOD). Bucci et al.~\citep{bucci2020ROS} focus on the effectiveness of image rotation for open set domain adaptation, and propose rotation-based open set (ROS), which use the rotation-invariance as a metric to identify the private class samples. Jang et al.~\cite{jang2022UADAL} propose unknown-aware domain adversarial learning (UADAL) to align the source and the target-shared distribution while segregating the target-private distribution. Both progressive graph learning (PGL)~\citep{luo2020PGL} and manifold regularized joint transfer (MRJT)~\citep{liu2023MRJT} essentially set empirical thresholds to assign pseudo-labels to confidence samples from the target domain. Huang et al.~\cite{Huang24BCL} propose a correlation metric-based graph framework which employs Hilbert-Schmidt independence criterion to characterize the separation between unknown and known classes. Adjustment and alignment (ANNA)~\citep{li2023ANNA} aims to improve the two-step pipeline using causality-based debiasing. However, its applicability is limited to OSDA scenarios and does not extend to PDA. By contrast, we view OSDA and PDA as cases of extreme label shift and address both scenarios.  

\subsection{Partial Domain Adaptation} 

PDA assumes that the label space of the target domain is a subset of the source domain, thus, it needs to identify the private class in the source domain and then mitigate the effect of negative transfer. 

 Pioneering PDA works are devoted to reweighing samples based on adversarial networks. Cao et al. propose partial adversarial domain adaptation (PADA)~\cite{cao2018PADA} which adds class-level weights to the source domain classifier. Technically, it weights both the classification loss and the adversarial loss on the source domain samples. Zhang et al.~\cite{zhang2018IWAN} propose the importance weighted adversarial nets (IWAN), which have separate feature extractors for each domain, and then view the domain discrimination as a bi-class problem. Subsequently, Cao et al.~\cite{cao2019ETN} propose the example transfer network (ETN) which introduces an additional classifier and a progressive weighting scheme based on the transferability of source domain samples. Chen et al.~\cite{chen2022DARL} introduce domain adversarial learning to learn the shared feature subspace of selected source domain instances and target domain instances. Sahoo et al.~\cite{sahoo2023SLM} introduce a selector network and make instance-level binary predictions to select the source shared class samples for subsequent adversarial domain alignment.
 
 Unlike methods based on adversarial learning to solve partial domain adaptation problems, Li et al.~\cite{li2020DRCN} propose the deep residual correction network (DRCN) to address optimization difficulties and cross-domain differences using MMD. Thopalli et al.~\cite{thopalli2023DeepOPA} propose deep orthogonal Procrustes alignment (DeepOPA), which combines deep features with orthogonal Procrustes alignment to achieve effective subspace alignment for domain adaptation. Note that the accuracy of target domain prediction plays an important role in estimating the class-wise weights in the source domain. Liang et al.~\cite{liang2020BA3US} use samples in the source domain to make data augment for the target domain. Yang et al.~\cite{yang2023CLA} propose a new reweighing scheme by incorporating the weights generated by the target domain information. Tian et al.~\cite{fang2023PLSC} follow the alignment and separation assumptions to train an SVM for pseudo-labeling the target samples for private class identification. As mentioned above, these methods often rely on unrealistic assumptions for successful private class identification, limiting their applicability in real-world scenarios.

\subsection{Optimal Transport for Domain Adaptation}

Optimal transport (OT) has achieved great success in domain alignment to learn invariant representation~\cite{luo2021ckb}. Courty et al.~\cite{courty2017otda} introduce OT for domain adaptation to align the source and target marginal distributions. Li et al.~\cite{li2020etd} further incorporate the label information into the cost matrix to gain more precise transport. Since marginal alignment may fail to learn invariant representation space when domain discrepancy is large, joint distribution optimal transport (JDOT)~\cite{courty2017jdot} is proposed for joint distribution alignment. To improve JDOT and estimate an unbiased transport plan when using mini-batches, OT with relaxed marginal constraints~\cite{fatras2021uot,nguyen2022pot} are proposed. Another efficient way to learn invariant representation is conditional distribution alignment. To achieve conditional alignment, Ren et al.~\cite{ren2022buresnet} extend the optimal transport to the reproducing kernel Hilbert space. Wang et al.~\cite{wang2024PPOT} propose a probability-polarized optimal transport for conditional distribution alignment. 

Recently, OT with mask ~\citep{zhang2022maskWD} is proposed for knowledge transfer to achieve intra-class alignment, implicitly aligning the conditional distribution. Gu et al.~\citep{gu2022keypoint} further incorporate the mask mechanism into Gromov-Wasserstein. Note that the strict marginal constraints severely limit their applications in more realistic scenarios such as PDA. In light of this, masked optimal transport (MOT)~\citep{luo2023MOT} has been proposed by relaxing the strict marginal constraints. However, in more challenging OSDA scenarios, the theoretical understanding of masked OT has not yet been developed. Thus, it remains necessary to explore the theoretical foundation of masked OT for successful OSDA.

\section{Methodology}\label{sec3}

Under the settings of extreme label shift in domain adaptation, there is a labeled source domain $\mathcal{D}_s=\{\bm{x}_i^s,y_i^s\}_{i=1}^{n_s}$ and an unlabeled target domain $\mathcal{D}_t=\{\bm{x}_j^t\}_{j=1}^{n_t}$ drawn from different distributions $P\neq Q$. Notably, for any $\bm{x}_j^t$, its true label $y_j^t$ is also deterministic though unavailable. Moreover, let $\mathcal{Y}_s$ and $\mathcal{Y}_t$ represent their label spaces, respectively. Then $\mathcal{Y}_s$ and $\mathcal{Y}_t$ are heterogeneous, e.g., 
\begin{itemize}
    \item In OSDA, $\mathcal{D}_t$ contains private class data, i.e., $\mathcal{Y}_s\subsetneq \mathcal{Y}_t$. 
    \item In PDA, $\mathcal{D}_s$ contains private class data, i.e., $\mathcal{Y}_s\supsetneq \mathcal{Y}_t$.
\end{itemize}

Our final goal is to learn a classification model that classifies target samples as correct as possible, where target private classes not in $\mathcal{Y}_s\cap\mathcal{Y}_t$ will be uniformly considered as one $private$ class. A model $h\circ g$ consists of a feature transformation $g$ and a classifier $h$. The transformation $g$ outputs a feature vector $\bm{z}=g(\bm{x})$ as the representation of $\bm{x}$. Classifier $h$ receives the representation as input and outputs the predicted label $\hat{y} =h\circ g(\bm{x})$.

\subsection{Preliminary}
Optimal transport focuses on finding an optimal solution for moving mass from one distribution to another under constraints. We will first provide a brief introduction to OT. 

We denote by $\mathcal{S}_n$ the set of histograms with $n$ bins: $\mathcal{S}_n=\left\{S\in \mathbb{R}^n_{+}, S^{\top}\bm{1}_n=1\right\}$, where $\bm{1}_n$ denotes the all-ones vector in $\mathbb{R}^n$. Let $P_X\in \mathcal{S}_n$ and $P_{X'}\in \mathcal{S}_m$ be two empirical distributions over two sample sets $\{\bm{x}_i\}_{i=1}^{n}$ and $\{\bm{x}_j'\}_{j=1}^{m}$, i.e., $P_{X}= \frac{1}{n}\sum_{i}\delta_{\bm{x}_{i}}$ and $P_{X'}= \frac{1}{m}\sum_{j} \delta_{\bm{x}_{j}'}$. Then the set of all possible couplings between $P_X$ and $P_{X'}$ is given by 
\begin{align*}
    \Pi(P_X,P_{X'})=\left\{\bm{\Gamma} \in \mathbb{R}_{+}^{n \times m} \mid \bm{\Gamma} \bm{1}_{m}=P_X, \bm{\Gamma}^{\top} \bm{1}_{n}=P_{X'}\right\}, 
\end{align*}
where $\bm{\Gamma}$ is a transport plan with the entry $\Gamma_{ij}$ that describes the amount of mass transported from $\bm{x}_i$ to $\bm{x}_j'$. Now given a cost matrix $\bm{C}\in \mathbb{R}^{n\times m}$ that measures the distance between $\{\bm{x}_i\}_{i=1}^{n}$ and $\{\bm{x}_j'\}_{j=1}^{m}$. With the entropy regularization $H(\bm{\Gamma})=\sum_{i,j}\Gamma_{ij}\ln \Gamma_{ij}$, the classical entropy regularized OT between $P_X$ and $P_{X'}$ is written as~\cite{Sinkhorn-dist13,courty2017otda} 
\begin{align*}
    \mathrm{OT}^{\lambda}(P_X,P_{X'},\bm{C})=\argmin\limits_{\bm{\Gamma} \in \Pi(P_X,P_{X'})}\langle\bm{\Gamma}, \bm{C}\rangle_{F} + \lambda H(\bm{\Gamma}),
\end{align*}
where $\lambda>0$ is the parameter of sparsity penalty, and $\langle\bm{\Gamma},\bm{C}\rangle_{F}=\sum_{ij} \Gamma_{ij} C_{ij}$ is the Frobenius inner product.  

Such an optimization objective seeks an sample-level matching which usually ignores and even degrades the discriminability of the transport plan. It is also limited in handling category shift challenges due to the hard constraints. By replacing the strict constraints with relaxed penalty terms and introducing the mask mechanism, Luo et al.~\citep{luo2023MOT} propose masked optimal transport to address the problems above. Suppose labels $\{{y}_i\}_{i=1}^{n}$ and $\{{y}_j'\}_{j=1}^{m}$ are known\footnote{The labels may be not completely known in the unsupervised domain adaptation tasks, thus, they are estimated by pseudo labels.}, the mask matrix $\bm{M}\in \mathbb{R}^{n\times m}$ is defined as 
\begin{equation*}
    M_{ij}:= \left\{
    \begin{aligned}
    1,\,&      & {\rm if} \,\, {y}_i={y}_j',\\
    \infty,&      & {\rm if} \,\, {y}_i\neq{y}_j'.\\
    \end{aligned} \right.
\end{equation*}
Then the masked cost matrix is defined as $\bar{\bm{C}}=\bm{C}\odot \bm{M}$, and the masked OT is formulated as
\begin{align*}
    \mathrm{MOT}^{\lambda}_{\beta_1,\beta_2}(P&_X,P_{X'},\bar{\bm{C}})=\argmin\limits_{\bm{\Gamma} \in \mathbb{R}_{+}^{n \times m}}~\langle\bm{\Gamma}, \bar{\bm{C}}\rangle_{F} + \lambda H(\bm{\Gamma})\\
    &+\beta_1 D_{{\rm KL}}(\bm{\Gamma} \bm{1}_{n}\|P_X)+\beta_2 D_{{\rm KL}}(\bm{\Gamma}^{\top} \bm{1}_{m}\|P_{X'}),
\end{align*}
where $D_{{\rm KL}}$ is the Kullback-Leibler Divergence, $\beta_1$ and $\beta_2$ are non-negative penalty parameters. 

As aforementioned, on the whole data manifold,  the variance within a single shared class can be significantly larger than the difference between a private class and another shared class. However, in a local neighborhood of a shared class cluster, the distance between two samples with shared class is always smaller than that between shared and private class samples. We will formulate this key observation as the local spatial structure of data. Then MOT is used to leverage this local structure and establish a general method for identifying the private class.  

\subsection{Locality-aware Private Class identification}
\label{subsec:private_identification}

The private class identification problems in extreme label shift can be uniformly described as follows. Given two sample sets $\bm{X}_1=\{\bm{x}_i\}_{i=1}^{n}$ and $\bm{X}_2=\{\bm{x}_j'\}_{j=1}^{m}$, they share $K$ classes while $\bm{X}_2$ contains extra private class samples\footnote{For simplicity, we do not distinguish which is the source domain and which is the target domain here.}. We aim to identify the private class samples in $\bm{X}_2$.
 
\begin{figure}[t] \centering
    \includegraphics[scale=0.65]{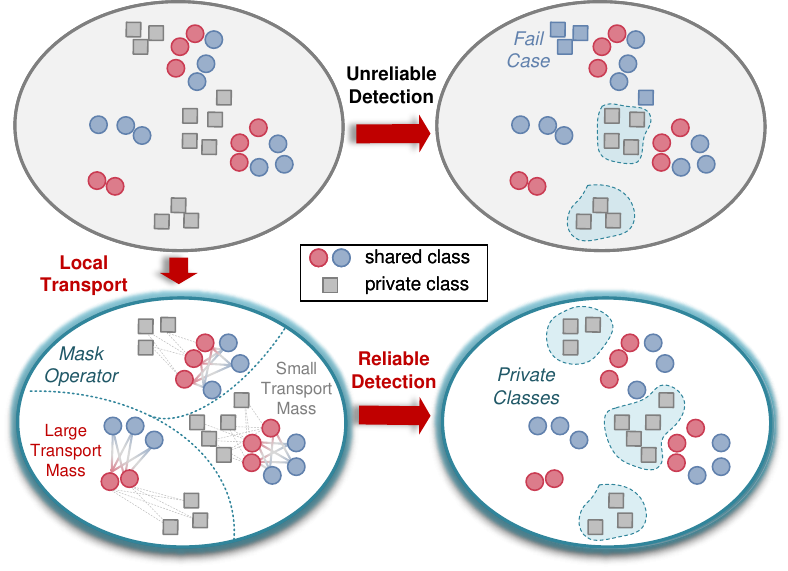}\hspace{0in}
    \caption{Illustration of most existing private class identification methods (top), and our OT-based reliable private class identification method (down). Better view in color.}
    \label{fig:spatial_structure_for_identification}
\end{figure}

In our method, the magnitude relationship between shared and private shifts is formulated as the local spatial structure, which is further characterized by masked OT. The identification is finally achieved through the magnitude relationship between transported mass. As shown in Figure~\ref{fig:spatial_structure_for_identification}, the method relaxes the strong assumptions on the whole data manifold structure, achieving more reliable identification than previous methods in practical applications. 



\label{sec:private class identification}

Specifically, we will formulate the local spatial structure as the magnitude relationship of the expected distances between instances with the same label. Given labels $\{\bar{y}_i\}_{i=1}^{n}$ and $\{\bar{y}_j'\}_{j=1}^{m}$ corresponding to $\bm{X}_1$ and $\bm{X}_2$, respectively. We divide the representations $g(\bm{X}_2)=\{g(\bm{x}_j')\}_{j=1}^{m}$ into small subsets $\mathcal{F}_k=\{\bm{z}_j'\in g(\bm{X}_2)\mid \bar{y}_{j}'=k\}$. Within each subset $\mathcal{F}_k$, the representations corresponding to the private and shared classes are denoted by $\mathcal{F}_k^{prv}=\{\bm{z}_j'\in\mathcal{F}_k\mid y_{j}'=K+1\}$ and $\mathcal{F}_k\backslash\mathcal{F}_k^{prv}$, respectively. Here $y_{j}'$ is the true label of $\bm{z}_j'$ and $K+1$ represents the $private$. Essentially, $\mathcal{F}_k^{prv}$ denotes a set of private class representations with label $\bar{y}_{j}'=k$, and $\mathcal{F}_k\backslash\mathcal{F}_k^{prv}$ represents a set of shared class representations with label $\bar{y}_{j}'=k$. In addition, let $P_{Z^{prv}_k}$ and $P_{Z^{shr}_k}$ denote two empirical distributions over $\mathcal{F}^{prv}_k$ and $\mathcal{F}_k\backslash\mathcal{F}_k^{prv}$, respectively. Then, given a non-negative function $f$ as the distance metric, the local spatial structure is reformulated as the magnitude relationship between the distances, i.e.,
\begin{align}
\label{eq:distance_property}
    \mathbb{E}_{P_{Z^{shr}_k}}[f(\bm{z},\bm{z}')]<\mathbb{E}_{P_{Z^{prv}_k}}[f(\bm{z},\bm{z}')],
\end{align}
where $k \in [\![K]\!]$,  and $\bm{z}$ is any representation belongs to $ \{\bm{z}_i\in g(\bm{X}_1)\mid \bar{y}_{i}=k\}$. In other words, when considering representations that share the same label, the expected distance between two such shared class representations is always smaller than that between a shared class representation and one from a private class.

\subsubsection{OT-based Local Spatial Structure Characterization}\hfill

Masked OT will be built, and the formulated local structure in Eq.~\eqref{eq:distance_property} will be converted to the magnitude relationship of entries in the transport plan.
 
Let $\bm{C}^{Z}\in \mathbb{R}^{n\times m}$ denote the transport cost between $g(\bm{X}_1)$ and $g(\bm{X}_2)$, with entries given by $C^{Z}_{ij}=c(g(\bm{x}_i),g(\bm{x}_j'))$ for a given cost function $c(\cdot,\cdot)$ (e.g., the squared Euclidean distance). Let $\bm{M}\in \mathbb{R}^{n\times m}$ be the mask matrix estimated with labels $\{\bar{y}_i\}_{i=1}^{n}$ and $\{\bar{y}_j'\}_{j=1}^{m}$, $\bar{\bm{C}}^{Z}=\bm{C}^{Z}\odot \bm{M}$ be the masked cost matrix, and $(P_Z, P_{Z'})\in S_{n}\times S_{m}$ represent two empirical distributions over $g(\bm{X}_1)$ and $g(\bm{X}_2)$, respectively. The optimal transport plan $\bm{\Gamma}^{*}\in \mathbb{R}^{n\times m}$ is 
\begin{align}
\label{eq:gamma_computation}    \bm{\Gamma}^{*}=\mathrm{MOT}^{\lambda}_{\infty,\beta_2}(P_Z,&P_{Z'},\bar{\bm{C}}^{Z}).
\end{align}
Here $\beta_1$ is set to $\infty$, which means a unilateral strict marginal constraint on $\bm{\Gamma}^{*}$, i.e., $\bm{\Gamma}^{*} \bm{1}_{m}=P_Z= \frac{1}{n}\sum_{i}\delta_{\bm{z}_{i}}$. The constraint ensures that $\sum_{j=1}^{m} \Gamma^{*}_{ij}=\frac{1}{n}$, i.e., each representation in $g(\bm{X}_1)$ transports an equal amount of mass $\frac{1}{n}$. This implies each representation within $g(\bm{X}_1)$ is equally important in the transportation. An intuitive explanation is that all representations in $g(\bm{X}_1)$ originate from the shared classes, and successful private class identification only relies on the information from the shared class instances. Accordingly, every representation within $g(\bm{X}_1)$  wields considerable influence on the process of private class identification. It is necessary to acknowledge that each representation plays an equally important role.

It can be proved that the optimal transport plan $\bm{\Gamma}^{*}$ has the following properties.
\begin{prop}\label{prop:MOT_property}
(distance-related inequality). Given $\bm{z}_i\in  g(\bm{X}_1)$ with $\bar{y}_i=k$, and $\bm{z}_j'$, $\bm{z}_l'\in \mathcal{F}_k\subset g(\bm{X}_2)$. If the cost function $c(\cdot,\cdot)$ satisfies $c(\bm{z}_i,\bm{z}_j')<c(\bm{z}_i,\bm{z}_l')$, then $\exists\,\lambda>0$ and sufficiently
small $\beta_2>0$, s.t., $\bm{\Gamma}^{*}$ satisfies
    \begin{equation*}
        \Gamma^{*}_{ij}>\Gamma^{*}_{il}.
    \end{equation*}
\end{prop}


The proof is placed on the supplementary material. Proposition~\ref{prop:MOT_property} tells us that $\bm{\Gamma}^*$ tends to assign larger values to representation pairs with smaller distances. Intuitively, if the local spatial structure holds, then for $\bm{z}_j'\in\mathcal{F}_k\backslash\mathcal{F}_k^{prv}$ and $\bm{z}_l'\in\mathcal{F}_k^{prv}$, the amount of mass transported from those $\bm{z}_i$ with $\bar{y}_i=k$ to $\bm{z}_j'$ is expected to be larger than that to $\bm{z}_l'$, i.e.,

\begin{align*}
    \sum_{\bar{y}_i=k} \Gamma^{*}_{ij}>\sum_{\bar{y}_i=k} \Gamma^{*}_{il}.
\end{align*}

According to the local transport property of masked OT~\cite{luo2023MOT}, $\Gamma^{*}_{ij}=0$ if $\bar{y}_i\neq \bar{y}_j$, thus the above inequality is equivalent to
\begin{align}
\label{eq:kno_greater_unk}
    \sum_{i=1}^{n} \Gamma^{*}_{ij}>\sum_{i=1}^{n} \Gamma^{*}_{il},
\end{align}
i.e., the local spatial structure can be described by the magnitude relationship between the total mass transported to shared and private class samples.

\subsubsection{Private Class Identification through Transport Mass}

Inspired by the above discussion, a private class identification method can be naturally induced utilizing the transport mass.

Equation~\eqref{eq:kno_greater_unk} indicates that the mass transported to a private class representation tends to be small due to the local structure. In other words, by reversing the sign of transport mass, an identification score $s(\bm{z}_j')$ for measuring the likelihood that $\bm{z}_j'$ belongs to private can be defined as
\begin{align}
\label{eq:score_def}
    s(\bm{z}_j') :=\frac{1}{m}-\sum_{i=1}^{n} \Gamma^{*}_{ij}.
\end{align}

\begin{figure}[t] \centering
    \includegraphics[scale=0.7]{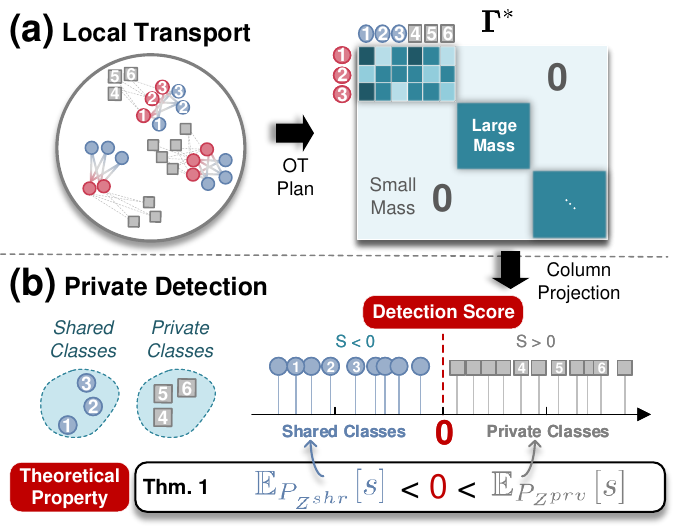}\hspace{0in}
    \caption{Illustration of Theorem~\ref{th:expectation_magnitude_relations} on guiding the identification of private samples. The local spatial structure is characterized by $\bm{\Gamma}^{*}$, then identification score function $s$ acts on columns of $\bm{\Gamma}^{*}$ and we can identify shared and private samples based on the values of $s$. }
    \label{fig:method_illustration}
\end{figure}

It is obvious that the mean of the scores over $g(\bm{X}_2)$ equals to zero. So intuitively, if $\bm{z}_j'$ belongs to private class, then $s(\bm{z}_j')>0$; otherwise $s(\bm{z}_j')<0$. Let $\mathcal{F}^{prv}=\bigcup\limits_{k=1}^{K+1}\mathcal{F}_k^{prv} \subset g(\bm{X}_2)$ denote all private class representations within $g(\bm{X}_2)$, and $\mathcal{F}^{shr}=\bigcup\limits_{k=1}^{K+1}(\mathcal{F}_k\backslash\mathcal{F}_k^{prv})\subset g(\bm{X}_2)$ denote all shared class representations in $g(\bm{X}_2)$. $P_{Z^{prv}}$ and $P_{Z^{shr}}$ represent the empirical distributions over $\mathcal{F}^{prv} $ and $\mathcal{F}^{shr} $, respectively. Then we have the following theoretical result.
\begin{theorem}
\label{th:expectation_magnitude_relations}
    If the inequality~\eqref{eq:distance_property} holds for $f(\bm{z},\bm{z}')=\exp(c(\bm{z},\bm{z}'))$ and $\forall\,k \in [\![K]\!]$, then $\exists\,\lambda>0 $ and sufficiently small $\beta_2>0$, s.t.,
    \begin{align*}
    \mathbb{E}_{P_{Z^{shr}}}[s(\bm{z}')]<0<\mathbb{E}_{P_{Z^{prv}}}[s(\bm{z}')].
    \end{align*}
\end{theorem}


The proof is placed on the supplementary material. Theorem~\ref{th:expectation_magnitude_relations} highlights the score function’s strong ability to distinguish between shared and private class samples. Specifically, it ensures that under the local spatial structure assumption, the expectation scores of shared and private class representations are strictly separated by 0. This further implies that zero can serve as an effective threshold for private class identification, i.e., we classify representations with $s>0$ into private as presented in Figure~\ref{fig:method_illustration}. 

Although zero can serve as a threshold for identification, its application is contingent upon certain assumptions as described in Theorem~\ref{th:expectation_magnitude_relations}. To broaden the applicability of our method in various practical scenarios, we would relax these assumptions by introducing a margin. This margin would ensure that only samples with significant scores are considered private. To induce an appropriate margin, we rethink $\bm{\Gamma}^*$ from the insight of joint distribution and consider the following probability matrix $\bm{\Gamma}'\in \mathbb{R}^{K\times m}$, i.e.,
\begin{align*}
    \bm{\Gamma}':=\begin{bmatrix}
\sum_{y_i=1} \Gamma^{*}_{i1} & \cdots & \sum_{y_i=1} \Gamma^{*}_{im}\\
\vdots & \ddots & \vdots \\
\sum_{y_i=K} \Gamma^{*}_{i1} & \cdots & \sum_{y_i=K} \Gamma^{*}_{im}
\end{bmatrix}.
\end{align*}

Let $P_{Y}$ be the empirical label distribution determined by $\bm{X}_1$, i.e., $P_{Y}= \frac{1}{n}\sum_{i}\delta_{y_{i}}$, where $y_{i}\in[\![K]\!]$ is the true label of $\bm{x}_i$. Then $\bm{\Gamma}'$ satisfies $\bm{\Gamma}'\bm{1}_{m}=P_{Y}$ and $ \bm{\Gamma}'^{\top} \bm{1}_{K}\approx P_{Z'}$, so it can be viewed as an approximate probability coupling between $P_{Y}$ and $P_{Z'}$. Although $\Gamma'_{kj}$ is not an exact joint probability, it provides a meaningful characterization of the co-occurrence at $(Y=k, Z'=\bm{z}_j')$ in the sense of optimal transport. Larger values indicate stronger correspondence between $\bm{z}_j'$ and class $k$, Thus, $\Gamma'_{kj}$ can be intuitively understood as an analog of the probability that $\bm{z}_j'$ corresponds to class $k$. Since $[\![K]\!]$ consists of shared classes, we further interpret $\sum_{k=1}^{K}\Gamma'_{kj}=\sum_{i=1}^{n} \Gamma^{*}_{ij}$ as an analog of the probability that $\bm{z}_j'$ corresponds to shared classes. Recall that $s(\bm{z}_j')$ measures the likelihood that $\bm{z}_j'$ belongs to the private class. Intuitively, for a private class sample, its score $s$ should exceed its probability of belonging to the shared classes. This leads to the inequality $s(\bm{z}_j')=\frac{1}{m}-\sum_{i=1}^{n} \Gamma^{*}_{ij}>\sum_{k=1}^{K}\Gamma'_{kj}=\sum_{i=1}^{n} \Gamma^{*}_{ij}$ by the definition of $s$ in Eq.~\eqref{eq:score_def}, which yields $s(\bm{z}_j')>\frac{1}{2m}$ for a private class sample $\bm{z}_j'$.

Consequently, we can obtain the set of potential shared class samples $\mathcal{D}^{shr}\subset \bm{X}_2$ with $s<0$, and the set of potential private class samples $\mathcal{D}^{prv}\subset \bm{X}_2$ with $s>\frac{1}{2m}$, i.e., 
\begin{equation}
\label{eq:private_class_identification}
\begin{aligned}
    &\mathcal{D}^{shr}=\left\{\bm{x}_j'\in \bm{X}_2 \mid s(g(\bm{x}_j'))<0\right\}, \\ 
    &\mathcal{D}^{prv}=\left\{\bm{x}_j' \in \bm{X}_2 \mid s(g(\bm{x}_j'))>1/2m\right\}.
\end{aligned}
\end{equation}
The private class identification is finally achieved by obtaining these two sets. 

We summarize the entire process of private class identification in extreme label shift scenarios as follows. For practical DA tasks with extreme label shift, it is essential to have prior knowledge to ascertain which set, either $\{\bm{x}_i^s\}_{i=1}^{n_s}$ or $\{\bm{x}_j^t\}_{j=1}^{n_t}$, contains the private class samples. We will designate the set containing the private class samples as $\bm{X}_2$ and assign the remaining set to $\bm{X}_1$. The corresponding labels $\{\bar{y}_i\}_{i=1}^{n}$ and $\{\bar{y}_j'\}_{j=1}^{m}$ will be determined by true/predicted labels, depending on the availability of the true labels; if the true labels are not available, the predicted labels will be used. Subsequently, we will built masked OT from $\bm{X}_1$ to $\bm{X}_2$ and calculate the optimal transport plan $\bm{\Gamma}^{*}$ by Eq.~\eqref{eq:gamma_computation}. Leveraging $\bm{\Gamma}^{*}$, we can then identify the shared class samples \(\mathcal{D}^{shr}\) and the private class samples \(\mathcal{D}^{prv}\) using Eq.~\eqref{eq:private_class_identification}.


\subsection{Algorithm and Applications for Extreme Label Shift}\label{ReOT}
Based on the locality-aware private class identification method, we now
present the numerical algorithm for solving DA under two extreme label shift scenarios, i.e., open set domain adaptation and partial domain adaptation.


The algorithm can be roughly divided into two steps. First, the private and shared class samples will be identified. Then, knowledge transfer will be implemented. 

For convenience, w.l.o.g., we consider the case $\mathcal{Y}_s\subsetneq \mathcal{Y}_t$, i.e., OSDA. The results can be easily extended to $\mathcal{Y}_s\supsetneq \mathcal{Y}_t$, i.e., PDA. The condition $\mathcal{Y}_s\subsetneq \mathcal{Y}_t$ signifies that target domain $\{\bm{x}_j^t\}_{j=1}^{n_t}$ contains private class samples. Therefore, as mentioned in Sec.~\ref{subsec:private_identification}, we substitute sets $\bm{X}_1$ and $\bm{X}_2$ with $\{\bm{x}_i^s\}_{i=1}^{n_s}$ and $\{\bm{x}_j^t\}_{j=1}^{n_t}$, respectively. The corresponding labels $\{\bar{y}_i\}_{i=1}^{n}$ and $\{\bar{y}_j'\}_{j=1}^{m}$ are given by true labels $\{y_i^s\}_{i=1}^{n_s}$ and predicted labels $\{\hat{y}_i^t\}_{j=1}^{n_t}$, respectively. The default cost function $c$ is given by the squared Euclidean distance. The optimal transport plan $\bm{\Gamma}^{*}$ is then computed by Eq.~\eqref{eq:gamma_computation}, and we obtain the identified target shared class samples $\mathcal{D}^{shr}$ and the identified target private class samples $\mathcal{D}^{prv}$ by Eq.~\eqref{eq:private_class_identification}. These identified samples form the foundation for constructing a reliable transfer model in subsequent steps.

\subsubsection{Reliable Transfer Across Domains}\hfill

To ensure correct classification of these samples, we need to consider aligning the shared class-conditional distributions across domains through the cross-domain intra-class alignment on shared classes, which can be achieved by utilizing the local transport property of masked OT. 

Specifically, we approximate the labels of target samples by the predicted labels, and then build a masked OT from source domain to target domain by using samples in the shared classes with their labels. The cost function $c$ is given by the squared Euclidean distance. Notably, in the case where $\mathcal{Y}_s\subsetneq \mathcal{Y}_t$, the $\bm{\Gamma}^{*}$ computed during private class identification is served as a masked optimal transport plan from the source to the target domain. By setting the mass transported between shared and private classes in $\bm{\Gamma}^{*}$ to zero, we obtain the desired transport plan $\Gamma^{shr}_{ij}$ from the source shared to target shared classes, i.e.,
\begin{equation*}
    \Gamma^{shr}_{ij}:= \left\{
    \begin{aligned}
    0,\,\,\,\quad      & {\rm both} \,\, \bm{x}_i^s\,\operatorname{and}\,\bm{x}_j^t\notin \mathcal{D}^{shr},\\
    \Gamma^{*}_{ij},\quad      & \operatorname{others}.\\
    \end{aligned} \right.
\end{equation*}
 Then, the alignment term seeks $g$ to minimize the cross-domain intra-class distances, formulated as
\begin{align*}
    \min_{g} ~\langle\bm{\Gamma}^{shr}, \bm{C}^{Z}\rangle_{F},
\end{align*}
where $\bm{C}^{Z}$ measure the square Euclidean distance between $\{\bm{z}_i^s\}_{i=1}^{n_s}$ and $\{\bm{z}_j^t\}_{j=1}^{n_t}$. 

In addition, to enhance the model discrimination ability and avoid mismatching between the shared and private class samples, we also consider enlarging the discrepancy between the shared and private classes in the representation space. First, similar to $\bm{\Gamma}^{shr}$, a transport plan $\bm{\Gamma}^{prv}$ between shared class samples and private class samples is defined by
\begin{equation*}
    \Gamma^{prv}_{ij}:= \left\{
    \begin{aligned}
    0,\,\,\,\quad      & {\rm both} \,\, \bm{x}_i^s\,\operatorname{and}\,\bm{x}_j^t\notin \mathcal{D}^{prv},\\
    \Gamma^{*}_{ij},\quad      & \operatorname{others}.\\
    \end{aligned} \right.
\end{equation*}

Similarly, we use $\bm{C}^{Z}$ to denote the square Euclidean distance between $\{\bm{z}_i^s\}_{i=1}^{n_s}$ and $\{\bm{z}_j^t\}_{j=1}^{n_t}$, then the separation term is induced by maximizing the distances between shared and private class samples, which can be equivalently reformulated as the following objective, i.e.,
\begin{align*}
\min_{g}~-\langle\bm{\Gamma}^{prv}, \bm{C}^{Z}\rangle_{F}.
\end{align*}



Finally, the reliable transfer loss $\mathcal{L}_{rt}$ is obtained by combining alignment term and separation term:
\begin{align}
\label{eq:rt_loss}
    \min_{g}\mathcal{L}_{rt}= \langle\bm{\Gamma}^{shr}, \bm{C}^{Z}\rangle_{F}-\langle\bm{\Gamma}^{prv}, \bm{C}^{Z}\rangle_{F}.
\end{align}

\subsubsection{Risk Minimization}\hfill

To ensure basic model performance, the empirical classification risk $\mathcal{L}_{cls}$ is minimized. Specifically, it minimizes the risk not only for shared classes but also for the private. In fact, if the model can precisely recognize the private samples, it can also determine which samples belong to the shared classes. Therefore, the model performance on shared classes can benefit from the risk minimization on the private. Let $\ell(\cdot,\cdot)$ represent a given loss function, $\mathcal{L}_{cls}$ is written as 
\begin{align}
\label{eq:cls_loss}
    \min_{g,h}\mathcal{L}_{cls}=\frac{1}{|\mathcal{D}_s\cup\mathcal{D}^{prv}|}{\mathlarger{\sum}}_{\bm{x}\in\mathcal{D}_s\cup\mathcal{D}^{prv}} \ell( h\circ g(\bm{x}),y),
\end{align}
where pseudo label $y=|\mathcal{Y}_s\cap\mathcal{Y}_t|+1$ is used for all $\bm{x}\in\mathcal{D}^{prv}$, since $\mathcal{D}^{prv}\subset\mathcal{D}_t$ and the true labels are unavailable.


Additionally, we propose to learn risk on the transported source domain, which is also called barycenter reconstruction loss. Specifically, we map the target representations back to the source domain based on the transport plan $\bm{\Gamma}^{*}$ and then minimize the reconstruction loss. The mapping is established via the barycenter map problem:
\begin{align*}
    \hat{\bm{z}}_i^s= \argmin_{z} \sum_{j=1}^{n_t}\Gamma^{*}_{ij}c(\bm{z},\bm{z}^t_j).
\end{align*}
As demonstrated by Courty et al.~\citep{courty2017otda}, an analytic solution for this problem can be written as
\begin{align*}
    \hat{\bm{z}}_i^s= n_s \sum_{j=1}^{n_t}\Gamma^{*}_{ij}\bm{z}^t_j.
\end{align*}
The barycenter reconstruction loss is then formulated as
\begin{align}
\label{eq:barycenter_loss}
    \min_{g,h}\mathcal{L}_{br} = \frac{1}{n_s} \sum_{i=1}^{n_s} \ell(h(\hat{\bm{z}}_i^s),y_i^s).
\end{align}

The barycenter mapping leverages target samples in the representation space to reconstruct the source sample $\bm{z}^s$ with minimal cost. By minimizing $\mathcal{L}_{br}$, each target sample receives soft supervision proportional to its contribution to the reconstruction. The target samples that belong to the same class as $\bm{z}^s$ will contribute more significantly. Consequently, this soft supervision will focus on pushing the target samples close to their corresponding class prototypes, implicitly lowering their classification risk. Additionally, this process enhances the representation space and accentuates the local spatial structure, thereby improving the accuracy of private class identification. 

Let $\eta_1, \eta_2>0$ be trade-off parameters, our final training objective is written as
\begin{align}\label{final_obj}
    \min_{g,h} \mathcal{L}=\mathcal{L}_{cls}+\eta_1 \mathcal{L}_{rt}+\eta_2 \mathcal{L}_{br}.
\end{align}


\begin{algorithm}[!t]
    \caption{ReOT for extreme label shift DA}
    \begin{algorithmic}[1]\label{Alg:ReOT}
        \renewcommand{\algorithmicrequire}{\textbf{Input:}}
        \renewcommand{\algorithmicensure}{\textbf{Output:}}
        \REQUIRE source domain data $\{\bm{x}_i^s,y_i^s\}_{i=1}^{n_s}$, target domain data $\{\bm{x}_j^t\}_{j=1}^{n_t}$, parameters $\lambda$, $\beta_2$, $\eta_1$ and $\eta_2$, number of iterations $T$;
        \ENSURE  classifier $h(\cdot)$ and feature transformation $g(\cdot)$;\\
        \STATE Initialize $h(\cdot)$ and $g(\cdot)$; 
        \FOR {$i = 1$ to $T$}
        \STATE Forward propagate $\{\bm{x}_j^t\}_{j=1}^{n_t}$ to obtain$\{\hat{y}_j^t\}_{j=1}^{n_t}$;
        \CASE {$\mathcal{Y}_s\subsetneq\mathcal{Y}_t$} 
        \STATE  Substitution: $\bm{X}_1\leftarrow \{\bm{x}_i^s\}_{i=1}^{n_s}$, $\bm{X}_2\leftarrow \{\bm{x}_j^t\}_{j=1}^{n_t}$; and
        $\{\bar{y}_i\}_{i=1}^{n}\leftarrow\{y_i^s\}_{i=1}^{n_s}$, $\{\bar{y}_j'\}_{j=1}^{m}\leftarrow\{\hat{y}_j^t\}_{j=1}^{n_t}$; 
        \STATE  Compute $\bm{\Gamma}^{*}$ by Eq.~\eqref{eq:gamma_computation}; 
        \STATE  Implement private class identification by Eq.~\eqref{eq:private_class_identification};
        \ENDCASE
        
        \CASE {$\mathcal{Y}_s\supsetneq\mathcal{Y}_t$} 
        \STATE Substitution: $\bm{X}_1\leftarrow \{\bm{x}_j^t\}_{j=1}^{n_t}$, $\bm{X}_2\leftarrow \{\bm{x}_i^s\}_{i=1}^{n_s}$; and $\{\bar{y}_i\}_{i=1}^{n}\leftarrow\{\hat{y}_j^t\}_{j=1}^{n_t}$, $\{\bar{y}_j'\}_{j=1}^{m}\leftarrow\{y_i^s\}_{i=1}^{n_s}$; 
        \STATE  Compute $\bm{\Gamma}^{*}$ by Eq.~\eqref{eq:gamma_computation}; 
        \STATE  Implement private class identification by Eq.~\eqref{eq:private_class_identification};
        \STATE  $\bm{\Gamma}^{*}=(\bm{\Gamma}^{*})^T$; \% $\bm{\Gamma}^{*}$ is a transport plan from target to source domain
        \ENDCASE
        \STATE Compute $\mathcal{L}_{rt}$, $\mathcal{L}_{cls}$, $\mathcal{L}_{br}$ by Eqs.~\eqref{eq:rt_loss},~\eqref{eq:cls_loss},~\eqref{eq:barycenter_loss}, respectively;
        \STATE Update $h(\cdot)$ and $g(\cdot)$ by Eq.~\eqref{final_obj}.
        \ENDFOR
    \end{algorithmic}
\end{algorithm}


For OSDA, it satisfies $\mathcal{Y}_s\subsetneq \mathcal{Y}_t$, so the above discussion can be directly applied. Besides, as neural network (NN) generally gains better learning capacity, we parameterize the classifier $h(\cdot)$ and feature transformation $g(\cdot)$ using NNs. The NN-based model can be updated with batch gradient descent. The overall training algorithm for OSDA is summarized in Algorithm~\ref{Alg:ReOT}, corresponding to case $\mathcal{Y}_s\subsetneq \mathcal{Y}_t$.

For PDA, since $\mathcal{Y}_s\supsetneq \mathcal{Y}_t$ and $\{\bm{x}_i^s\}_{i=1}^{n_s}$ contains private class samples, we just need to substitute $\bm{X}_1$ and $\bm{X}_2$ with $\{\bm{x}_j^t\}_{j=1}^{n_t}$ and $\{\bm{x}_i^s\}_{i=1}^{n_s}$, while substituting $\{\bar{y}_i\}_{i=1}^{n}$ and $\{\bar{y}_j\}_{j=1}^{m}$ with $\{\hat{y}_j^t\}_{j=1}^{n_t}$ and $\{y_i^s\}_{i=1}^{n_s}$, respectively, for computing $\bm{\Gamma}^*$ in Eq.~\eqref{eq:gamma_computation}. Then the private class identification is also implemented by Eq.~\eqref{eq:private_class_identification}. Now, $\bm{\Gamma}^*$ transports probability mass from the target to the source domain, thus the subsequently reliable transfer cannot directly proceed. Fortunately, since optimal transport is symmetric, we can directly transpose matrix $\bm{\Gamma}^*$ to obtain the transport plan from source to target. Using the transposed $\bm{\Gamma}^*$, $\mathcal{L}_{rt}$, $\mathcal{L}_{cls}$, and $\mathcal{L}_{br}$ can also be computed by Eqs.~\eqref{eq:rt_loss},~\eqref{eq:cls_loss}, and~\eqref{eq:barycenter_loss}, respectively. Notably, $\mathcal{Y}_s\supsetneq \mathcal{Y}_t$ implies that private class samples belong to the labeled source domain, i.e., $\mathcal{D}^{prv}\subset\mathcal{D}_s$. Consequently, there is no need to use the pseudo label for $\mathcal{L}_{cls}$ computation. By instantiating $h(\cdot)$ and $g(\cdot)$ as NNs, the overall training algorithm for PDA is also summarized in Algorithm~\ref{Alg:ReOT}, corresponding to case $\mathcal{Y}_s\supsetneq \mathcal{Y}_t$.

\subsection{Generalization Error Analysis}

In this section, we study the interpretability and transferability of the model learned by ReOT from theoretical perspective. By decomposing the error gap between the source and target domains, theoretical bounds will be provided for the severe label shift scenarioes. 

The theoretical bounds for OSDA have been given by Fang et al.~\cite{fang2020DAOD,zhong2021bridging}, but it is not applicable for addressing PDA tasks. Zhao et al.~\cite{zhao2019learning,li2021learning} give sufficient upper bounds for domain adaptation, but handling the bounds in the extreme label shift scenario is challenging. Combes et al.~\cite{tachet2020GLS} provide the theoretical bounds for generalized label shift domain adaptation. However, the conclusions are formulated under the closed set scenario, i.e., the source domain and target domain have identical label space. Therefore, it is necessary to extend the conclusions to the extreme label shift scenarioes where label spaces are heterogeneous, including both OSDA and PDA.

Let $p(Y)$ and $q(Y)$ defined on $\mathcal{Y}:=\mathcal{Y}_s\cup\mathcal{Y}_t$ represent the probability mass of the true label distribution of $P$ and $Q$, respectively. For convenience, w.l.o.g, the source/target private classes not in $\mathcal{Y}_s\cap\mathcal{Y}_t$ are uniformly denoted by $K_s/K_t$, i.e., $\mathcal{Y}=(\mathcal{Y}_s\cap\mathcal{Y}_t)\cup K_s\cup K_t$ and $p(Y=K_t)=q(Y=K_s)=0$. Now let $P(\hat{Y}\mid Y)$ represent the conditional predict distribution given $P$ and a predictor $\hat{Y}:=h\circ g(X)$. Then the error of model $h\circ g$ on source domain can be defined as $ \varepsilon_s(h \circ g):=\sum_{i\neq j\in\mathcal{Y}} p(\hat{Y}=i, Y=j)$. The error on target domain $\varepsilon_t(h\circ g)$ is defined similarly. Before presenting the main results, we first introduce two
important concepts for theoretical analysis as Combes et
al.~\cite{tachet2020GLS}.   
\begin{definition}
    (Maximal Prediction Error). The maximal error of predictor $ \hat{Y} $ on source distribution $ P $ is
    \begin{align*}
        \Delta_{{\rm MPE}}(P)(\hat{Y}\|Y):=\max \limits_{j\in \mathcal{Y}_s\cap\mathcal{Y}_t} p(\hat{Y}\neq j\mid Y=j).
    \end{align*}
\end{definition}

Intuitively, maximal prediction error implies the maximum gap between true label $Y$ and predicted label $\hat{Y}$. Moreover, it can also be viewed as the degree of error for the prediction. If the predictor can correctly classify shared class samples drawn from the distribution $P$, then the maximal prediction error $\Delta_{{\rm MPE}}(P)(\hat{Y}\|Y)$ will be zero. 

\begin{definition}
    (Shared Conditional Gap). Given source distribution $P$ and target distribution $Q$, the conditional gap of predictor $ \hat{Y} $ on shared classes is
    \small
    \begin{align*}
        \begin{split}
            \Delta_{{\rm SCG}}(\hat{Y}):=\sum_{j\in\mathcal{Y}_s\cap\mathcal{Y}_t}q(Y=j)\mathbb{D}(P(\hat{Y}|Y=j),Q(\hat{Y}|Y=j)),
        \end{split}
    \end{align*}
    \normalsize
    where
    \begin{align*}
        \mathbb{D}(P(\hat{Y}|Y=j),Q(\hat{Y}|Y=j)):=
        &\max_{i\neq j,i\in\mathcal{Y}}|p(\hat{Y}=i|Y=j)\\
        &-q(\hat{Y}=i|Y=j)|.
    \end{align*}
\end{definition}

The shared conditional gap measures the cross-domain conditional discrepancy on shared classes. A smaller shared conditional gap implies a smaller conditional discrepancy across domains. In fact, as demonstrated by Combes et
al.~\cite{tachet2020GLS}, if the following process is a Markov chain:
\begin{equation*}
    X\stackrel{g}{\longrightarrow}Z\stackrel{h}{\longrightarrow}\hat{Y},
\end{equation*}
where $Z=g(X)$ and $\hat{Y}=h\circ g(X)$, then $ P(g(X)|Y=j)=Q(g(X)|Y=j) $ for any $j\in \mathcal{Y}_s\cap\mathcal{Y}_t$ implies the shared conditional gap $\Delta_{{\rm SCG}}(\hat{Y})$ is zero. In other words, when the cross-domain class-conditional distributions on shared classes are aligned and $ \hat{Y} $ depends only on $ g(X) $, the shared conditional gap $\Delta_{{\rm SCG}}(\hat{Y})$ will be zero.

Based on these two concepts, we present our result on the generalization error in the following theorem.

\begin{theorem}\label{th:th2}
Given source distribution $P$ and target distribution $Q$, then for any predictor $\hat{Y}=h\circ g(X)$, 
\begin{equation*}
\label{eq:th2}
\begin{aligned}
    &|\varepsilon_t(h\circ g)-\varepsilon_{s}(h\circ g)|\leq\\
    &\|p(Y)-q(Y)\|_1\cdot \Delta_{{\rm MPE}}(P)(\hat{Y}\|Y) + (|\mathcal{Y}|-1) \Delta_{{\rm SCG}}(\hat{Y})\\
    &+p(\hat{Y}\neq Y, Y=K_s)+q(\hat{Y}\neq Y,Y=K_t),
\end{aligned}
\end{equation*}
where
\begin{align*}
    \|p(Y)-q(Y)\|_1=\sum_{i\in\mathcal{Y}_s\cap\mathcal{Y}_t}\left|p(Y=i)-q(Y=i)\right|.
\end{align*}
\end{theorem}

The proof is placed on the supplementary material. The upper bound in Theorem \ref{th:th2} provides a new way to decompose the error gap between source and target domains for the heterogeneous label space scenario, and it can also be directly used to obtain a generalization upper bound of the target risk $ \varepsilon_t $, i.e.,
\begin{equation*}
    \begin{aligned}
        \varepsilon_t&\leq \varepsilon_{s}+p(\hat{Y}\neq Y,Y=K_s)\\
        &+\|p(Y)-q(Y)\|_1\cdot \Delta_{{\rm MPE}}(P)(\hat{Y}\|Y)\\
        &+ (|\mathcal{Y}|-1) \Delta_{{\rm SCG}}(\hat{Y}) \\
    &+q(\hat{Y}\neq Y,Y=K_t).
    \end{aligned}
\end{equation*}

Note that $p(\hat{Y}\neq Y,Y=K_s)\leq \varepsilon_{s}$, thus it can be simplified as
\begin{equation}\label{eq:error_bound}
    \begin{aligned}
        \varepsilon_t&\leq 2\varepsilon_{s}+\|p(Y)-q(Y)\|_1\cdot \Delta_{{\rm MPE}}(P)(\hat{Y}\|Y)\\
        &+ (|\mathcal{Y}|-1) \Delta_{{\rm SCG}}(\hat{Y})+q(\hat{Y}\neq Y,Y=K_t).
    \end{aligned}
\end{equation}


This generalization upper bound, shown by the right-hand-side of inequality~\eqref{eq:error_bound}, consists of four terms. The first term measures the classification error of model $h \circ g$ on the source domain and can be sufficiently minimized by ReOT, as $\mathcal{L}_{cls}$ reduces the risk on the source domain.


The second term contains $\Delta_{{\rm MPE}}(P)(\hat{Y}\|Y)$ and $\|p(Y)-q(Y)\|_1$. As mentioned, $\Delta_{{\rm MPE}}(P)(\hat{Y}\|Y)$ will decrease to $0$ as the model achieves correct classification on source shared classes. Meanwhile, $\|p(Y)-q(Y)\|_1$ measures the discrepancy between the true label distributions and is a constant. Therefore, the second term can also be minimized by ReOT through risk minimization on the source domain.

The third term involves constant $|\mathcal{Y}|-1$ and the shared conditional gap $\Delta_{{\rm SCG}}(\hat{Y})$. By optimizing $\mathcal{L}_{rt}$, ReOT decreases the intra-class distances on shared classes, and thus mitigates the cross-domain class-conditional discrepancy on shared classes. Our experiments will show that $\mathcal{L}_{rt}$ is indeed effective in alleviating class-conditional discrepancy. Thus, the third term can be also reduced by ReOT.


The fourth term measures the classification error on target private class samples. In PDA, it is naturally $0$ since there is no target private class. In OSDA, this term can be reduced by the risk minimization on target private class samples. Although the target private class samples are unlabeled in OSDA, most of them can be recognized by the proposed private class identification method. Consequently, ReOT can decrease this term by using $\mathcal{L}_{cls}$ for risk minimization on the private class samples. 

In conclusion, our ReOT can sufficiently decrease the upper bound of the target risk, thus achieving a small generalization error on the target domain. This ensures the effectiveness of ReOT from theoretical perspective. 

\section{Experiments}\label{sec4}

\subsection{Datasets and Setup}
\textbf{Datasets.}
We evaluate ReOT on four benchmark datasets, i.e., Image-CLEF \cite{caputo2014imageclef}, Office-31 \cite{saenko2010office31}, Office-Home \cite{venkateswara2017officehome}, and VisDA-2017~\cite{peng2017visda}. We follow the
standard protocols~\cite{cao2018PADA,gu2021AR,liu2023MRJT,li2023ANNA} to generate adaptation tasks. Details for datasets and settings are summarized as follows.
\begin{itemize}
    \item \textbf{Image-CLEF} consists of 12 categories from 4 domains, i.e., Caltech(C), ImageNet (I), Pascal (P), and Bing (B), that are collected from existing public datasets. Each domain contains 600 images. For the OSDA setting, the first 6 categories in alphabetical order are selected as shared classes and the remaining 6 are selected as target private classes. For the PDA setting, the target domain consists of the first 6 categories (in alphabetical order).
    \item \textbf{Office-31} contains 4652 images of 31 categories. The images are collected from 3 different domains, i.e., Amazon (A), DSLR (D), and Webcom (W). For the OSDA setting, the first 10 categories in alphabetical order are regarded as the shared class and the last 11 categories are considered target private classes. Following the protocols in~\cite{cao2018PADA,gu2021AR}, the target domain
consists of 10 categories in the PDA setting.
    \item \textbf{Office-Home} contains 15500 images from 4 domains, including Art (Ar), Clipart (Cl), Product (Pr), and Real-World (Rw), each domain contains 65 categories. For the OSDA setting, the first 25 categories (in alphabetical order) are selected as shared classes, while the rest are selected as target private classes. For the PDA setting, the first 25 classes (in alphabetical order) are selected as the target domains.
    \item \textbf{VisDA-2017} contains about 200k images of 12 categories from 2 domains, i.e., Synthetic (S) and Real (R). The first 6 classes (in alphabetical order) are taken as the target domain in the PDA setting. The PDA task is considered as the knowledge transfer from S to R.
\end{itemize}

\textbf{Evaluation Metrics.}
In the OSDA setting, we use three evaluation metrics following the mainstream OSDA works~\citep{liu2023MRJT,li2023ANNA}, i.e., OS*, UNK, and H. OS* is the average classification accuracy over shared classes and UNK is the classification accuracy of the private class. ${\rm H}=2\times\frac{{\rm OS^*}\times{\rm UNK}}{{\rm OS^*}+{\rm UNK}}$ is the harmonic mean accuracy of OS* and UNK, and it is the core metric since a high H appears only if the model performs well on both OS* and UNK. In the PDA setting, we use the standard classification accuracy as evaluation metrics following the mainstream PDA works~\cite{cao2018PADA,gu2021AR}.

\textbf{Implementation.}
We employ ResNet-50~\cite{he2016resnet} pre-trained on ImageNet~\cite{deng2009imagenet} as the backbone, which has been widely used in OSDA and PDA, then we fix it to extract the 2048-dimensional feature vectors as input for all datasets. The feature transformation $ g(\cdot) $ consists of two fully-connected-layers and the setting for dimensions is $\mathbb{R}^{2048} \to \mathbb{R}^{1024} \to \mathbb{R}^{256}$. The classifier $ h(\cdot) $ consists of a single fully-connected-layer with $ |\mathcal{Y}_s|+1 $-dimensional output and SoftMax activation. The loss function $\ell(\cdot,\cdot)$ is defined as the cross-entropy between the classifier's output and the one-hot encoded form of the true label. The final predicted label is determined by the index of the maximum component value. $g(\cdot)$ and $h(\cdot)$ are initially fine-tuned with the empirical risk on the source domain. Then the whole model $h\circ g$ is optimized by Algorithm~\ref{Alg:ReOT}. More details about the training implementations are provided in the supplementary material.




\subsection{Comparative Experiments Under OSDA Setting}


  
        
        
        
        

 
        


\begin{table*}[!t]
\small
\renewcommand{\arraystretch}{1.2}
 \renewcommand\tabcolsep{0.21pc}
    \begin{center}
    \caption{Classification accuracies (\%) in the OSDA setting on Image-CLEF dataset. Bold fonts indicate the best results, and the rest are the same.}\label{tab:CLEF}
    \vspace{-0.8pc}
    \begin{tabular}{c|ccc|ccc|ccc|ccc|ccc|ccc|ccc}
        \toprule
        \multicolumn{4}{c|}{\multirow{2}{*}{\textbf{Image-CLEF}}} & \multicolumn{3}{c|}{B$\rightarrow$C} & \multicolumn{3}{c|}{B$\rightarrow$I} & \multicolumn{3}{c|}{B$\rightarrow$P}& \multicolumn{3}{c|}{C$\rightarrow$B}& \multicolumn{3}{c|}{C$\rightarrow$I}& \multicolumn{3}{c}{C$\rightarrow$P}\\

        \multicolumn{4}{c|}{}&OS* &UNK &H &OS* &UNK &H & OS* &UNK &H &OS* &UNK &H &OS* &UNK &H &OS*&UNK&H\\
        \hline
 
        \multicolumn{4}{c|}{OSBP~\citep{saito2018OSBP}}
        & 87.0 & 81.0 & 83.9    & 85.3 & 65.7 & 74.3    & 66.3 & 66.7 & 66.5 
        & \textbf{62.0} & 58.0 & 59.9    & 89.0 & 80.0 & 84.3    & \textbf{87.7} & 53.7 & 66.7 \\
        
        \multicolumn{4}{c|}{STA~\citep{liu2019STA}}
        & 93.3 & 51.7 & 66.5    & 86.0 & 60.7 & 71.2    &\textbf{77.7} & 48.7 & 59.8 
        & 61.3 & 69.7 & 65.2    & \textbf{91.7} & 66.7 & 77.2    & 84.0 & 54.0 & 65.7 \\
        
        \multicolumn{4}{c|}{ROS~\citep{bucci2020ROS}}
        & 78.3 & 90.0 & 83.8    & 73.0 & 76.3 & 74.6    & 59.0 & 67.3 & 62.9 
        & 59.0 & 68.3 & 63.3    & 78.3 & 83.0 & 80.6    & 68.7 & 78.7 & 73.3 \\

        \multicolumn{4}{c|}{DAOD~\citep{fang2020DAOD}}
        & 79.4 & 82.0 & 80.7    & 78.4 & 90.9 & 84.3    & 72.1 & 80.8 & 76.3 
        & 51.3 & 47.1 & 49.1    & 79.0 & 88.6 & 83.6    & 74.5 & 78.9 & 76.7 \\

        \multicolumn{4}{c|}{MRJT~\citep{liu2023MRJT}} 
         & 90.7 & \textbf{99.0} & 94.7   & 74.7 & \textbf{95.3} & 83.7    & 63.3 & \textbf{92.7} & 75.2 
         & 58.0 & 85.3 & 69.1   & 87.0 & 93.3 & 90.1    & 69.7 & 85.7 & 76.8 \\
        
        \multicolumn{4}{c|}{ANNA~\citep{li2023ANNA}} 
         & \textbf{95.3} & 98.3 & \textbf{96.8}   & 81.3 & 84.7 & 83.0    & 74.0 & 75.0 & 74.5 
         & 58.0 & 83.0 & 68.3   & 87.0 & 93.0 & 89.9    & 78.7 & 84.0 & 81.2 \\

        \rowcolor[gray]{.9}
        \multicolumn{4}{c|}{\textbf{ReOT}} 
        &93.2&95.8&94.5
	  &\textbf{90.0}&93.9&\textbf{91.9}
	&77.2&82.1&\textbf{79.6}
        &56.6&\textbf{90.0}&\textbf{69.4}
	  &88.9&\textbf{97.0}&\textbf{92.8}
        &78.1&\textbf{87.4}&\textbf{82.5}\\

        \midrule
        \multicolumn{1}{c|}{\multirow{2}{*}{-}} & \multicolumn{3}{c|}{I$\rightarrow$B}& \multicolumn{3}{c|}{I$\rightarrow$C} & \multicolumn{3}{c|}{I$\rightarrow$P} & \multicolumn{3}{c|}{P$\rightarrow$B}& \multicolumn{3}{c|}{P$\rightarrow$C}& \multicolumn{3}{c|}{P$\rightarrow$I}& \multicolumn{3}{c}{\textbf{Mean}}\\
        &OS* &UNK &H &OS* &UNK &H &OS* &UNK &H & OS* &UNK &H &OS* &UNK &H &OS* &UNK &H &OS*&UNK&H\\
        \hline
 
        \multicolumn{1}{c|}{OSBP}
         & 55.7 & 60.7 & 58.1   & 80.7 & 92.7 & 86.3    & 66.3 & 74.3 & 70.1 
         & 52.3 & 61.0 & 56.3   & 94.0 & 68.0 & 78.9    & 66.0 & 80.7 & 72.6 
         & 74.4 & 70.2 & 71.5 \\

        \multicolumn{1}{c|}{STA}
        & \textbf{62.3} & 54.0 & 57.9    & 94.0 & 53.7 & 68.4    & 80.7 & 59.0 & 68.2 
        & \textbf{61.3} & 43.7 & 51.0    & 93.7 & 47.7 & 63.2    & \textbf{90.0} & 51.0 & 65.1 
        & \textbf{81.3} & 55.1 & 65.0 \\

        \multicolumn{1}{c|}{ROS}
        & 58.0 & 59.7 & 58.8    & 88.7 & 92.7 & 90.6    & 78.0 & 76.0 & 77.0 
        & 47.3 & 59.3 & 52.7    & 71.3 & 90.3 & 79.7    & 79.7 & 81.3 & 80.5 
        & 69.9 & 76.9 & 73.1 \\

        \multicolumn{1}{c|}{DAOD}
        & 54.5 & 56.9 & 55.7    & 80.3 & 82.0 & 81.2    & 73.3 & 80.8 & 76.9 
        & 51.7 & 51.0 & 51.3    & 79.0 & 82.0 & 80.5    & 79.6 & 86.6 & 83.9 
        & 71.1 & 75.8 & 73.3 \\

        \multicolumn{1}{c|}{MRJT}
        & 56.7 & 83.3 & 67.5    & \textbf{97.3} & 81.3 & 88.6    & 79.0 & \textbf{88.7} & 83.5 
        & 52.0 & \textbf{86.3} & 64.9    & 93.7 & 80.0 & 86.3    & 87.3 & 87.7 & 87.5 
        & 75.7 & 88.2 & 80.7 \\
        
        \multicolumn{1}{c|}{ANNA}
        & 56.0 & 78.0 & 65.2    & 94.3 & 97.7 & \textbf{96.0}    & 80.7 & 82.7 & 81.7 
        & 54.0 & 73.7 & 62.3    & 94.0 & 93.7 & 93.8    & 85.0 & 83.3 & 84.2 
        & 78.2 & 85.6 & 81.4 \\
         
        \rowcolor[gray]{.9}
        \multicolumn{1}{c|}{\textbf{ReOT}} 

        &58.4&\textbf{87.6}&\textbf{70.1}
	    &93.2&\textbf{98.4}&95.8
	    &\textbf{81.7}&86.4&\textbf{83.9}
        &55.1&80.6&\textbf{65.4}
	    &\textbf{94.8}&\textbf{95.7}&\textbf{95.2}
	    &89.0&\textbf{95.4}&\textbf{92.1}
        &79.7&\textbf{90.9}&\textbf{84.4}\\
        \bottomrule
    \end{tabular}
    \end{center}
\end{table*}

\begin{table*}[t]
\small
\renewcommand{\arraystretch}{1.2}
 \renewcommand\tabcolsep{0.5pc}
    \begin{center}
    \caption{Ablation study results ($\%$) on different modules. }
        \label{tab:ablation}
        \vspace{-0.8pc}
 \renewcommand\tabcolsep{0.8pc}
    \begin{tabular}{ccc|cc|cc|cc|cc}
        \toprule
        \multicolumn{3}{c|}{Modules}  & \multicolumn{2}{c|}{Image-CLEF}&\multicolumn{2}{c|}{Office-31}&\multicolumn{2}{c|}{Office-Home}&\multicolumn{2}{c}{\textbf{Mean}}\\
            $\mathcal{L}_{cls}$&$\mathcal{L}_{rt}$&$\mathcal{L}_{br}$&OSDA&PDA&OSDA&PDA &OSDA&PDA&OSDA&PDA 	 \\
        \hline
        \ding{51}&\ding{55}&\ding{55} 

        &80.9&84.1
        &88.5&87.1
        &63.6&61.4
        &77.7&77.5\\
 
        \ding{51}&\ding{51}&\ding{55} 
         &83.2&92.2
        &91.1 &95.9
        &65.8&69.3
        &80.0&85.8\\
 
        \ding{51}&\ding{55}&\ding{51} 

        &82.1&86.9
        &90.6 &92.3   
        &66.4&68.8
        &79.7&82.7\\
 
        \ding{51}&\ding{51}&\ding{51} 
        
        &\textbf{84.4}&\textbf{93.4}
        &\textbf{93.4} &\textbf{98.4}
        &\textbf{70.3}&\textbf{78.7}
        &\textbf{82.7} &\textbf{90.2}\\
        \bottomrule
        \end{tabular}
    \end{center}
\end{table*}

\begin{table*}[!t]
\small
\renewcommand{\arraystretch}{1.2}
 \renewcommand\tabcolsep{0.164pc}
    \begin{center}
    \caption{Classification accuracies ($ \% $) in the OSDA setting on Office-31 dataset. }
        \label{tab:31}
        \vspace{-0.8pc}
    \begin{tabular}{c|ccc|ccc|ccc|ccc|ccc|ccc|ccc}
        \toprule
        \multicolumn{1}{c|}{\multirow{2}{*}{\textbf{Office-31}}} & \multicolumn{3}{c|}{A$\rightarrow$W} & \multicolumn{3}{c|}{A$\rightarrow$D} & \multicolumn{3}{c|}{D$\rightarrow$A}& \multicolumn{3}{c|}{D$\rightarrow$W}& \multicolumn{3}{c|}{W$\rightarrow$A}& \multicolumn{3}{c|}{W$\rightarrow$D}& \multicolumn{3}{c}{\textbf{Mean}}\\

        \multicolumn{1}{c|}{}&OS* &UNK &H &OS* &UNK &H & OS* &UNK &H &OS* &UNK &H &OS* &UNK &H &OS*&UNK&H  &OS* &UNK &H \\
        \hline
  
        \multicolumn{1}{c|}{OSBP~\citep{saito2018OSBP}}
        & 86.8 & 79.2 & 82.7    & 90.5 & 75.5 & 82.4    & 76.1 & 72.3  & 75.1 & 97.7 & 96.7 & 97.2  & 73.0 & 74.4 & 73.7    & 99.1 & 84.2 & 91.1 
        & 87.2 & 80.4 & 83.7 \\
        
        \multicolumn{1}{c|}{STA~\citep{liu2019STA}}
        & 92.1 & 58.0 & 71.0    & \textbf{95.4} & 45.5 & 61.6    & \textbf{94.1} & 55.0 & 69.4 
        & 97.1 & 49.7 & 65.5    & \textbf{92.1} & 46.2 & 60.9    & 96.6 & 48.5 & 64.4 
        & \textbf{94.6} & 50.5 & 65.5 \\
        \multicolumn{1}{c|}{ROS~\citep{bucci2020ROS}}
        & 88.4 & 76.7 & 82.1    & 87.5 & 77.8 & 82.4    & 74.8 & 81.2 & 77.9 
        & 99.3 & 93.0 & 96.0    & 69.7 & 86.6 & 77.2    & \textbf{100} & 99.4 & \textbf{99.7} 
        & 86.6 & 85.8 & 85.9 \\

        \multicolumn{1}{c|}{cUADAL~\citep{jang2022UADAL}}
        & 85.5 &95.1 &90.1   &85.6 &90.4 &87.9   &74.2 &87.8 &80.5    
        &98.7 &97.7 &98.2   & 65.6 &87.8 &75.1   &99.3 &99.4 &99.4    
        &84.8 &93.0 &88.5\\

        \multicolumn{1}{c|}{MRJT~\citep{liu2023MRJT}} 
         & 90.4 & 89.6 & 90.0   & 90.2 & 89.7 & 90.0    & 92.1 & 90.0 & \textbf{91.0} 
         & 98.3 & \textbf{100} & 99.1    & 91.0 & 94.7 & \textbf{92.9}    & \textbf{100} & 93.1 & 96.4 
         & 93.7 & 92.9 & 93.3 \\
        
        \multicolumn{1}{c|}{ANNA~\citep{li2023ANNA}} 
         & 82.8 & 88.4 & 85.5   & 93.2 & 76.1 & 83.8    & 75.4 & 91.1 & 82.5 
         & \textbf{99.4} & 99.6 & \textbf{99.5}   & 76.0 & 87.9 & 81.6    & \textbf{100} & 96.8 & 98.4       & 87.8 & 90.0 & 88.6 \\
 
        \rowcolor[gray]{.9}
        \multicolumn{1}{c|}{\textbf{ReOT}} 
        &\textbf{93.4}&\textbf{98.5}&\textbf{95.9}
        &94.6&\textbf{95.9}&\textbf{95.2}
        &79.7&\textbf{93.1}&85.9
        &99.1&96.4&97.7
        &78.7&\textbf{95.7}&86.3     
        &98.7&\textbf{100}&99.3
        &90.7&\textbf{96.6}&\textbf{93.4}\\

        \bottomrule
    \end{tabular}

    \end{center}
\end{table*}

\begin{table*}[!t]
\small
\renewcommand{\arraystretch}{1.2}
    \renewcommand\tabcolsep{0.20pc}
    \begin{center}
    \caption{Classification accuracies ($\%$) in the OSDA setting on Office-Home dataset. }
        \label{tab:Home}
\vspace{-0.8pc}
    \begin{tabular}{c|ccc|ccc|ccc|ccc|ccc|ccc|ccc}
        \toprule
        \multicolumn{4}{c|}{\multirow{2}{*}{\textbf{Office-Home}}} & \multicolumn{3}{c|}{Ar$\rightarrow$Cl} & \multicolumn{3}{c|}{Ar$\rightarrow$Pr} & \multicolumn{3}{c|}{Ar$\rightarrow$Rw}& \multicolumn{3}{c|}{Cl$\rightarrow$Ar}& \multicolumn{3}{c|}{Cl$\rightarrow$Pr}& \multicolumn{3}{c}{Cl$\rightarrow$Rw}\\

        \multicolumn{4}{c|}{}&OS* &UNK &H &OS* &UNK &H & OS* &UNK &H &OS* &UNK &H &OS* &UNK &H &OS*&UNK&H\\
        \hline
 
        \multicolumn{4}{c|}{OSBP~\citep{saito2018OSBP}}
        &50.2 &61.1& 55.1   & 71.8& 59.8& 65.2  & 79.3& 67.5& 72.9
        &59.4 &70.3& 64.3   &67.0& 62.7& 64.7   & 72.0& 69.2& 70.6\\
        
        \multicolumn{4}{c|}{STA~\citep{liu2019STA}}
        &50.8& 63.4& 56.3   & 68.7& 59.7& 63.7  & 81.1& 50.5& 62.1
        & 53.0& 63.9& 57.9  & 61.4& 63.5& 62.5  & 69.8& 63.2& 66.3\\
        
        \multicolumn{4}{c|}{ROS~\citep{bucci2020ROS}}
        &50.6 &74.1& 60.1   & 68.4 &70.3 &69.3  &75.8& 77.2& 76.5 
        &53.6& 65.5& 58.9   & 59.8 &71.6& 65.2  & 65.3& 72.2& 68.6\\

        \multicolumn{4}{c|}{DAOD~\citep{fang2020DAOD}}
        & \textbf{72.6} & 51.8 & 60.5    & 55.3 & 57.9 & 56.6    & 78.2 & 62.6 & 69.5 
        & 59.1 & 61.7 & 60.4    & 70.8 & 52.6 & 60.4    & \textbf{77.8} & 57.0 & 65.8 \\

        \multicolumn{4}{c|}{PGL~\citep{luo2020PGL}}
        & 63.3 & 19.1 & 29.3    & \textbf{78.9} & 32.1 & 45.6    & \textbf{87.7} & 40.9 & 55.8 
        & \textbf{85.9} & 5.3 & 10.0     & \textbf{73.9} & 24.5 & 36.8    & 70.2 & 33.8 & 45.6 \\
        
        \multicolumn{4}{c|}{cUADAL~\citep{jang2022UADAL}}
        & 55.0 &75.6 &63.6   &69.4 &73.9 &71.6   &82.2 &73.3 &77.5
        & 53.8 &82.0 &\textbf{65.0}   &61.1 &\textbf{77.4} &68.3   &69.3 &76.3 &72.6\\

        \multicolumn{4}{c|}{MRJT~\citep{liu2023MRJT}} 
         & 49.7 & 75.3 & 59.9   & 69.3 & 77.1 & 73.0    & 78.2 & 73.4 & 76.1 
         & 52.3 & \textbf{82.6} & 64.1   & 59.0 & 76.1 & 66.5    & 67.4 & 79.5 & \textbf{73.0} \\
        
        \multicolumn{4}{c|}{ANNA~\citep{li2023ANNA}} 
         &61.4 &78.7 &\textbf{69.0}  &68.3 &\textbf{79.9} &73.7  &74.1 &79.7 &76.8 
         &58.0 &73.1 &64.7  &64.2 &73.6 &68.6  &66.9 &\textbf{80.2} &\textbf{73.0}\\
 
        \rowcolor[gray]{.9}
        \multicolumn{4}{c|}{\textbf{ReOT}} 
        &57.1&\textbf{78.9}&66.3
	  &72.7&76.0&\textbf{74.3}
	&77.4&\textbf{81.8}&\textbf{79.5}
        &59.0&69.6&63.9
	  &64.1&75.4&\textbf{69.3}
        &70.2&75.2&72.6
\\

        \midrule
        \multicolumn{1}{c|}{\multirow{2}{*}{-}} & \multicolumn{3}{c|}{Pr$\rightarrow$Ar}& \multicolumn{3}{c|}{Pr$\rightarrow$Cl} & \multicolumn{3}{c|}{Pr$\rightarrow$Rw} & \multicolumn{3}{c|}{Rw$\rightarrow$Ar}& \multicolumn{3}{c|}{Rw$\rightarrow$Cl}& \multicolumn{3}{c|}{Rw$\rightarrow$Pr}& \multicolumn{3}{c}{\textbf{Mean}}\\
        &OS* &UNK &H &OS* &UNK &H &OS* &UNK &H & OS* &UNK &H &OS* &UNK &H &OS* &UNK &H &OS*&UNK&H\\
        \hline
 
        \multicolumn{1}{c|}{OSBP}
         &59.1 &68.1 &63.2  &44.5 &66.3 &53.2   &76.2 &71.7 &73.9 
         &66.1 &67.3 &66.7  &48.0 &63.0 &54.5   &76.3 &68.6 &72.3 
         &64.1 &66.3 &64.7\\

        \multicolumn{1}{c|}{STA}
        &55.4 &73.7 &63.1   &44.7 &71.5 &55.0   &78.1 &63.3 &69.7 
        &67.9 &62.3 &65.0   &51.4 &57.9 &54.2   &77.9 &58.0 &66.4 
        &63.4 &62.6 &61.9\\

        \multicolumn{1}{c|}{ROS}
        & 57.3 & 64.3 & 60.6    & 46.5 & 71.2 & 56.3    & 70.8 & 78.4 & 74.4 
        & 67.0 & 70.8 & 68.8    & 51.5 & 73.0 & 60.4    & 72.0 & 80.0 & 75.7 
        & 61.6 & 72.4 & 66.2 \\

        \multicolumn{1}{c|}{DAOD}
        & 71.3 & 50.5 & 59.1    & 58.4 & 42.8 & 49.4    & 81.8 & 50.6 & 62.5 
        & 66.7 & 43.3 & 52.5    & 60.0 & 36.6 & 45.5    & 84.1 & 34.7 & 49.1 
        & 69.6 & 50.2 & 57.6 \\

        \multicolumn{1}{c|}{PGL} 
         & \textbf{73.7} & 34.7 & 47.2   & \textbf{59.2} & 38.4 & 46.6    & \textbf{84.8} & 27.6 & 41.6 
         & \textbf{81.5} & 6.1 & 11.4    & \textbf{68.8} & 0.0 & 0.0      & \textbf{84.8} & 38.0 & 52.5 
         & \textbf{76.1} & 25.0 & 35.2 \\

        \multicolumn{1}{c|}{cUADAL}
        &50.9 &\textbf{82.4} &62.9   &41.2 &\textbf{80.7} &54.6   &71.2 &\textbf{83.4} &\textbf{76.8} 
        &66.8 &79.6 &\textbf{72.6}   &51.8 &71.1 &59.9   &77.8 &75.6 &76.7   
        &62.5 &\textbf{77.6 }&68.5\\
        
        \multicolumn{1}{c|}{MRJT}
        & 60.3 & 73.2 & 66.1    & 51.0 & 72.2 & 59.7    & 76.4 & 69.0 & 72.5 
        & 66.9 & 66.1 & 66.5    & 55.1 & 71.1 & 62.1    & 77.0 & 75.6 & 76.3 
        & 63.5 & 74.3 & 68.0 \\
        
        \multicolumn{1}{c|}{ANNA}
        & 63.0 & 70.3 & 66.5    & 54.6 & 74.8 & \textbf{63.1}    & 74.3 & 78.9 & 76.6 
        & 66.1 & 77.3 & 71.3    & 59.7 & 73.1 & \textbf{65.7}    & 76.4 & 81.0 & 78.7 
        & 65.6 & 76.7 & \textbf{70.7} \\
         
        \rowcolor[gray]{.9}
        \multicolumn{1}{c|}{\textbf{ReOT}} 

        &61.8&74.8&\textbf{67.7}
        &53.6&73.5&62.0
        &73.9&77.7&75.7
        &60.9&\textbf{80.7}&69.4
        &54.8&\textbf{77.1}&64.1
        &75.4&\textbf{84.7}&\textbf{79.7}
        
        &65.1&77.1&70.4
\\
        \bottomrule
        \end{tabular}
    \end{center}
\end{table*}

\begin{table*}[t]
    \small
    \renewcommand{\tabcolsep}{0.165pc} 
    \begin{center}
        \caption{Classification accuracies ($ \% $) in the PDA setting on Office-Home and VisDA-2017 datasets.}
        \label{tab:Home_VisDA_PDA}
        \vspace{-1.5pc}
        \begin{tabular}{c*{13}{c}c}
            \toprule
            \multirow{2}{*}{\textbf{Methods}} & \multicolumn{13}{c}{\textbf{Office-Home}} &
            \textbf{VisDA-2017}\\
            & Ar$ \to $Cl	& Ar$ \to $Pr	&Ar$ \to $Rw	&Cl$ \to $Ar	&Cl$ \to $Pr	&Cl$ \to $Rw	&Pr$ \to $Ar	&Pr$ \to $Cl	&Pr$ \to $Rw	&Rw$ \to $Ar	&Rw$ \to $Cl	&Rw$ \to $Pr& \textbf{Mean}	&S$ \to $R\\
            \midrule
            
            DANN \cite{ganin2016DANN}&43.8&67.9&77.5&63.7&59.0&67.6&56.8&37.1&76.4&69.2&44.3&77.5&61.7&51.0\\
            PADA \cite{cao2018PADA}&52.0&67.0&78.7&52.2&53.8&59.0&52.6&43.2&78.8&73.7&56.6&77.1&62.1&53.5
            \\
            IWAN~\cite{zhang2018IWAN} &53.9 &54.5 &78.1 &61.3 &48.0 &63.3 &54.2 &52.0 &81.3 &76.5 &56.8 &82.9 &63.6&-\\
            ETN \cite{cao2019ETN}&59.2&77.0&79.5&62.9&65.7&75.0&68.3&55.4&84.4&75.7&57.7&84.5&70.5&-\\
            HAFN \cite{xu2019AFN}&53.4&72.7&80.8&64.2&65.3&71.1&66.1&51.6&78.3&72.5&55.3&79.0&67.5&65.1\\
            SAFN \cite{xu2019AFN}&58.9&76.3&81.4&70.4&73.0&77.8&72.4&55.3&80.4&75.8&60.4&79.9&71.8&67.7\\
            BA$^3$US~\cite{liang2020BA3US}&60.6 &83.2 &88.4 &71.8 &72.8 &83.4 &75.5 &61.6 &86.5 &79.3 &62.8 &86.1 &76.0 &-\\
            DRCN~\cite{li2020DRCN} 
            &54.0 &76.4 &83.0 &62.1 &64.5 &71.0 &70.8 &49.8 &80.5 &77.5 &59.1 &79.9 &69.0 &58.2\\
            DMP \cite{luo2020dmp}&59.0 &81.2 &86.3 &68.1 &72.8 &78.8 &71.2 &57.6 &84.9 &77.3 &61.5 &82.9 &73.5 &72.7\\
            AR \cite{gu2021AR}& \textbf{67.4} &85.3 &90.0 & 77.3 &70.6 &85.2 &79.0 & \textbf{64.8} &89.5 &80.4 &66.2 &86.4 &78.3 &88.7\\
            m-POT \cite{nguyen2022pot}&64.6 &80.6 &87.2&\textbf{76.4}&77.6&83.6&77.1&63.7&87.6&81.4& \textbf{68.5}&87.4&78.0&87.0\\
            IDSP~\cite{li2022idsp}&60.8&80.8&87.3&69.3&76.0&80.2&74.7&59.2&85.3&77.8&61.3&85.7&74.9&-\\            SAN++~\cite{cao2022san}&61.3&81.6&88.6&72.8&76.4&81.9&74.5&57.7&87.2&79.7&63.8&86.1&76.0&63.1\\
            RAN~\cite{wu2022ran}&63.3&83.1&89.0&75.0&74.5&82.9&78.0&61.2&86.7&79.9&63.5&85.0&76.8&75.1\\
            MUL \cite{luo2022MUL}&57.4 & \textbf{88.7} &90.8 &71.0 &80.4 &82.1 &77.9 &59.8 &\textbf{91.2} & \textbf{83.5} &58.1 & 87.7 &77.4 &77.5\\
            DeepOPA
            \cite{thopalli2023DeepOPA}&51.1&65.9&76.4
            &65&65.76&72.8&64.2&49.3&74.6&70.4&54.9
            &77.3&65.6&-
            \\     SLM~\cite{sahoo2023SLM}&61.1&84.0&91.4&76.5&75.0&81.8&74.6&55.6&87.8&82.3&57.8&83.5&76.0& 91.7\\
            PLSC~\cite{fang2023PLSC}& 63.2	&85.7	&91.6	&72.1	&80.2	&82.7	&78.7	&56.1	&86.0	&76.6	&58.9	&87.6	&76.6
 &74.9\\

            \rowcolor[gray]{.9}\textbf{ReOT}    &63.7	&87.4	&\textbf{92.3}	&74.2	&\textbf{81.4}	&\textbf{87.1}	&\textbf{79.2}	&60.8	&88.0	&79.2	&62.8	&\textbf{88.5}	&\textbf{78.7}

 &\textbf{91.8}\\
            \bottomrule
        \end{tabular}
    \end{center}
\end{table*}

\begin{table}[!t]
\small
    \renewcommand{\arraystretch}{1.0}
    \begin{center}
        \caption{Classification accuracies ($ \% $) in the PDA setting on Image-CLEF dataset.}
        \label{tab:CLEF_PDA}
        \vspace{-1.5pc}
        \renewcommand{\tabcolsep}{0.35pc} 
        \begin{tabular}{c*{7}{c}}
            \toprule
            \textbf{Image-CLEF} &I$ \to $P	&P$ \to $I	&I$ \to $C	&C$ \to $I	&C$ \to $P	&P$ \to $C	& \textbf{Mean}	\\
            \midrule 
            DANN \cite{ganin2016DANN}&78.1&86.3&91.3&84.0&72.1&90.3&83.7\\
            PADA \cite{cao2018PADA}&81.7&92.1&94.6&89.8&77.7&94.1&88.3\\
            HAFN \cite{xu2019AFN}&79.1&87.7&93.7&90.3&77.8&94.7&87.2\\
            SAFN \cite{xu2019AFN}&79.5&90.7&93.0&90.3&77.8&94.0&87.5\\
            DMP \cite{luo2020dmp}&82.4 & 94.5 &96.7 &94.3 &78.7 &96.4 &90.5 \\
            MUL \cite{luo2022MUL}& 87.5 & 92.2 & 98.1 &94.6 &87.6 & 98.5 &93.1\\
            PLSC~\cite{fang2023PLSC}&85.0 &\textbf{97.3} &\textbf{98.3} &\textbf{96.3} &78.0 &\textbf{99.0} &92.3\\
            
            \rowcolor[gray]{.9}\textbf{ReOT} &\textbf{88.1}	&94.3	&98.1	&93.8	&\textbf{88.0}	&97.9	&\textbf{93.4}\\
            \bottomrule
        \end{tabular}
    \end{center}
\end{table}

\begin{table}[!t]
    \small
    \renewcommand{\arraystretch}{1.0}
    \begin{center}
        \caption{Classification accuracies ($ \% $) in the PDA setting on Office-31 dataset.}
        \label{tab:Office31_PDA}
        \vspace{-1.5pc}
        \renewcommand{\tabcolsep}{0.2pc} 
        \begin{tabular}{c*{7}{c}}
            \toprule
            \textbf{Office-31}&A$\to$W&D$\to$W&W$\to$D &A$\to$D&D$\to$A&W$\to$A
            &\textbf{Mean}	\\
            \midrule
            
            DANN~\cite{ganin2016DANN}&73.6&96.3&98.7&81.5&82.8&86.1&86.5\\
            PADA~\cite{cao2018PADA}&86.5&99.3&\textbf{100}&82.2&92.7&95.4&92.7\\
            IWAN~\cite{zhang2018IWAN} &89.2 &99.3 &99.4 &90.5 &95.6 &94.3 &94.7\\
            ETN~\cite{cao2019ETN}&94.5 &\textbf{100} &\textbf{100} &95.0 &96.2 &94.6 &96.7\\
            HAFN~\cite{xu2019AFN}&87.5&96.7&99.2&87.3&89.2&90.7&91.7\\
            SAFN~\cite{xu2019AFN}&87.5&96.6&99.4&89.8&92.6&92.7&93.1\\
            BA$^3$US~\cite{liang2020BA3US}&99.0 &\textbf{100} &98.7 &99.4 &94.8 &95.0 &97.8\\
            DRCN~\cite{li2020DRCN} &90.8 &\textbf{100} &\textbf{100} &94.3 &95.2 &94.8 &95.9\\
            
            DMP~\cite{luo2020dmp}&96.6 &\textbf{100} &\textbf{100} &96.4 &95.1 &95.4 &97.2 \\
            AR~\cite{gu2021AR}&93.5 &\textbf{100} &99.7&96.8&95.5&96.0&96.9\\
            IDSP~\cite{li2022idsp}&99.7&99.7&\textbf{100}&99.4&95.1&95.7&98.3\\
            SAN++~\cite{cao2022san}&99.7&\textbf{100}&\textbf{100}&98.1&94.1&95.5&97.9\\    
            RAN~\cite{wu2022ran}&99.0&\textbf{100}&\textbf{100}&97.7&\textbf{96.3}&96.2&98.2\\
            
            MUL~\cite{luo2022MUL}&94.2& \textbf{100}& \textbf{100}& 98.5& 95.6& 96.3& 97.5\\
            SLM~\cite{sahoo2023SLM}&\textbf{99.8}&\textbf{100}&99.8&98.7&96.1&95.9&98.4\\
            PLSC~\cite{fang2023PLSC}&99.3&\textbf{100}&\textbf{100}&\textbf{100}&96.0&\textbf{96.6}&\textbf{98.7}\\
            \rowcolor[gray]{.9}\textbf{ReOT} &99.3	&99.9	&\textbf{100}	&98.5	&96.1	&96.5	&98.4							
            \\
            \bottomrule
        \end{tabular}
    \end{center}
\end{table}

We focus more on H in performance evaluation, since the harmonic mean H is a composite metric of OS* and UNK. 

The results on Image-CLEF are presented in Table~\ref{tab:CLEF}.  From the view of OS*, we see some methods perform very well. For example, STA achieves the best OS* on 5 out of all 12 tasks and outperforms the ReOT method by 1.6$\%$ on average OS*. However, STA relies heavily on the empirical threshold to achieve private class identification, which lacks reliability and precision. As a result, it fails to identify the private samples effectively and gets significantly low UNK and H values. In contrast, ReOT utilizes a reliable private class identification approach, which ensures the precision of the identification. Consequently, the trained model can successfully recognize the private samples while not degenerating the performance over the shared classes. We see that ReOT achieves the best UNK on 7 out of all 12 tasks and outperforms the best baseline by 2.7$\%$ on average UNK (90.9$\%$), while maintaining a second-highest average of 79.7$\%$ OS*. There are several methods that also perform well on both OS* and UNK, e.g., MRJT and ANNA, which results in good performance in the view of H. Compared with these methods, ReOT achieves the best H on 10 out of all 12 tasks. Moreover, it gains the best average of 84.4$\%$ H over 12 OSDA tasks. It outperforms the state-of-the-art (SOTA) method ANNA by 1.5$\%$, 5.3$\%$ and 3.0$\%$ on average OS*, UNK and H, respectively.

Table~\ref{tab:31} presents the results on Office-31. The performance of ReOT on the Office-31 dataset also shows similar trends to the CLEF dataset. Specifically, ReOT achieves the best UNK on 5 out of all 6 tasks on Office-31 and outperforms the best baseline by 3.6\% on average UNK (96.6\%), while maintaining a high average of 90.7\% OS* value. Compared with other SOTA methods that cannot identify the private samples precisely, ReOT also gains the highest H value of 93.4\%. These results significantly verify the effectiveness of the proposed private class identification approach in helping the model correctly recognize the target private samples and improving model performance. 

Table~\ref{tab:Home} presents the results on Office-Home dataset. ReOT achieves an average H-value of 70.3\%, which is only lower than the causality-driven method ANNA by 0.4\% and outperforms all the other methods. Compared with related OSDA methods, such as MRJT which uses manifold regularization, ReOT improves the average OS*, UNK, and H values by 1.7\%, 2.5\%, and 2.3\%, respectively. All the above results demonstrate that our ReOT reaches a better balance in identifying the private samples and classifying the shared samples than SOTA methods. It outperforms SOTA models by an average of 2.0\%$\sim$2.5\% per dataset, significantly validating the stability and effectiveness in dealing with domain adaptation tasks with extreme label shift.

\subsection{Comparative Experiments Under PDA Setting}

Table~\ref{tab:CLEF_PDA} presents the results on Image-CLEF. In the sense of mean accuracy, ReOT significantly outperforms the SOTA methods. Specifically, by precise local structure characterization, ReOT improves the global manifold structure-based method DMP. It can be also observed that PLSC suffers a serious performance degradation on C$\to$P task, as it relies on the alignment and separation assumptions to train an SVM for private class identification. With relaxed assumptions, ReOT generally achieves the level of PLSC on all adaptation tasks while improving PLSC by 10$\%$ on the C$\to$P task.

Table~\ref{tab:Office31_PDA} presents the results on Office-31 dataset. In the PDA setting, tasks on Office-31 are relatively simple, i.e., the domain gap on shared classes is relatively small. Therefore, methods that rely on strong assumptions generally achieve high performance (close to 100$\%$). Compared with these methods, ReOT's mean accuracy is only slightly lower (0.3$\%$) than that of PLSC, yet it still ensures top performance among the other methods. This validates again the stability of ReOT facing different environments.

The results on Office-Home dataset are presented in Table~\ref{tab:Home_VisDA_PDA}. From the view of average accuracy, ReOT achieves the most advanced performance (78.7$\%$). Specifically, our ReOT achieves the best accuracy on 6 out of all 12 tasks. Compared with recent PDA methods that similarly implement private class identification (e.g., SLM and PLSC) under strong assumptions, ReOT improves them by an average of 2.4$\%$, demonstrating the reliability of the proposed method. 

The results on VisDA-2017 dataset are also presented in Table~\ref{tab:Home_VisDA_PDA}. VisDA-2017 is a more challenging dataset, containing larger-scale data and a more complicated cluster structure. It can be observed from the results that most existing methods have experienced serious performance degradation. For example, PLSC delivers the best performance on Office-31, but its results on VisDA-2017 are poor; MUL gains the second-best performance on Image-CLEF and comparable results on Office-Home, but it also suffers severe performance degradation on VisDA-2017. By contrast, ReOT still provides the most advanced performance on VisDA-2017. The overall results demonstrate that ReOT effectively maintains superior performance on challenging datasets, further validating the reliability and transferability of the model learned by ReOT. 

\begin{figure*}[!t] \centering
    \subfloat[W$\to$A (OSDA)]{
        \includegraphics[width=0.4\textwidth]{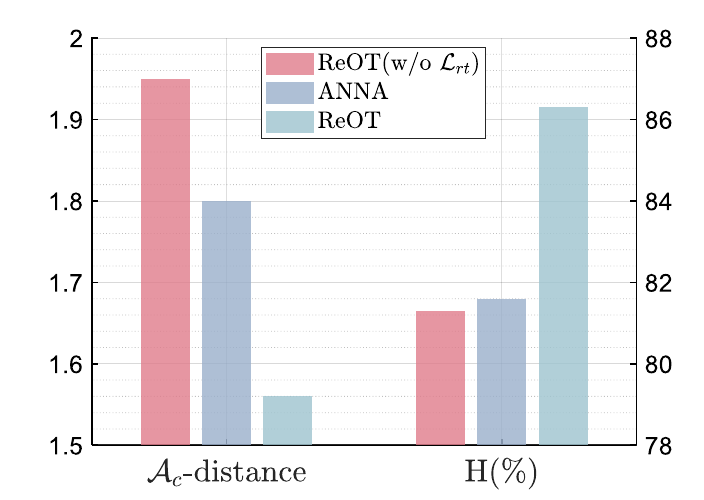}
        \label{fig:A_distance_OSDA}}\hspace{8mm}
    \subfloat[W$\to$A (PDA)]{
     \includegraphics[width=0.4\textwidth]{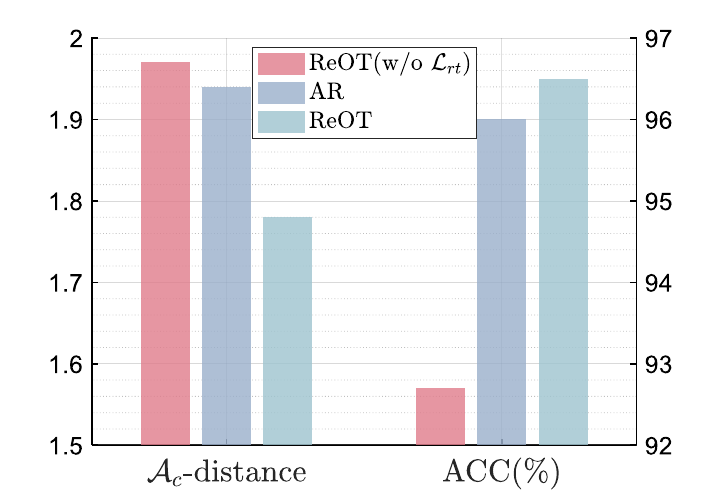}
        \label{fig:A_distance_PDA}}
    \caption{Comparison of cross-domain class-conditional discrepancies including $\mathcal{A}$-distance and H/ACC-value of different models. (a) OSDA. (b) PDA.}
    \label{fig:A_distance_bar}
\end{figure*}

\subsection{Other Analysis}
\textbf{Ablation Study.}
To evaluate the effectiveness of different modules, we conduct ablation experiments on three datasets. We present the harmonic accuracy (H) for OSDA and the standard accuracy (ACC) for PDA in Table~\ref{tab:ablation}.
It can be observed that, compared with the baseline model (Row 1), introducing the transfer module $\mathcal{L}_{rt}$ (Row 2) and the reconstruction module $\mathcal{L}_{br}$ (Row 3) both improve the H/ACC across all three datasets. Specifically, $\mathcal{L}_{rt}$ improves the mean H-value and ACC by 2.3$\%$ and 8.3$\%$, respectively, while $\mathcal{L}_{br}$ improves the mean H-value and ACC by 2.0$\%$ and 5.2$\%$, respectively. This highlights the benefit of each module in OSDA and PDA scenarios. Besides, introducing the transfer module shows a higher improvement on H-value and ACC over these datasets, indicating that the invariant representation learning via class-conditional alignment plays a crucial role in both OSDA and PDA. Furthermore, ReOT (Row 4) consistently achieves the best performance on all datasets, which implies that the two modules can supplement each other and significantly benefit the full model. In conclusion, the experimental results verify the general effectiveness of $\mathcal{L}_{rt}$ and $\mathcal{L}_{br}$ on different datasets.

\textbf{Class-conditional Discrepancy.}
To validate the effectiveness of ReOT with the alignment term $\mathcal{L}_{rt}$ for cross-domain class-conditional alignment on shared classes, we conduct experiments to analyze the conditional discrepancies of different models, i.e., ReOT (w/o $\mathcal{L}_{rt}$), ANNA, and ReOT. Here, ReOT (w/o $\mathcal{L}_{rt}$) represent the ReOT without the distribution alignment term $\mathcal{L}_{rt}$. For discrepancy measure, the $ \mathcal{A} $-distance \cite{ben2010theory} is applied to the class-wise data to estimate the conditional discrepancy on shared classes. The experiment results on Office-31 task W$ \to $A are presented in Figure~\ref{fig:A_distance_bar}, where the class-conditional $ \mathcal{A}_c$-distance is computed as the mean of $ \mathcal{A} $-distances on all class-wise data. From the results, we can observe in both OSDA and PDA, the domain discrepancy is large without alignment, and ReOT significantly mitigates the discrepancy by minimizing $\mathcal{L}_{rt}$. Specifically, as shown in Figure~\ref{fig:A_distance_OSDA}, in the OSDA setting, ANNA shows better capacity than ReOT (w/o $\mathcal{L}_{rt}$) in mitigating the domain discrepancy; however, by introducing $\mathcal{L}_{rt}$, ReOT gains a smaller $ \mathcal{A}_c$-distance with higher H-value, which implies that ReOT with $\mathcal{L}_{rt}$ learns a better invariant representation without degenerating the discriminative ability. A similar trend appears in the PDA setting as shown in Figure~\ref{fig:A_distance_PDA}. Overall, the results above validate that the proposed ReOT effectively decreases conditional discrepancy by optimizing $\mathcal{L}_{align}$.

\begin{figure*}[!t] \centering
    \subfloat[W$\to$A (OSDA)]{
        \includegraphics[width=0.235\textwidth]{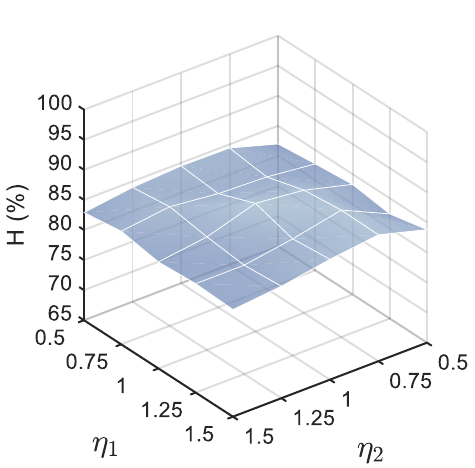}
        \label{fig:OSDA_sensitivity_31_WtoA}}
    \subfloat[Cl$\to$Pr (OSDA)]{
     \includegraphics[width=0.235\textwidth]{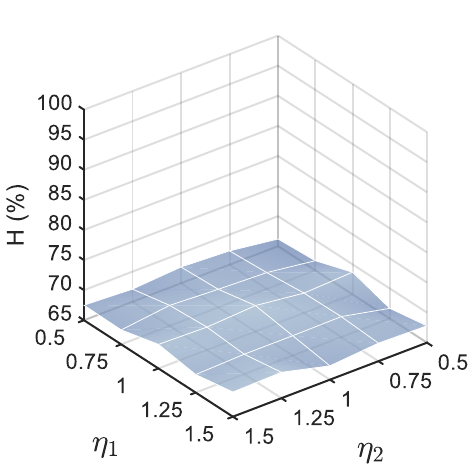}
        \label{fig:OSDA_sensitivity_Home_CLtoPR}}
    \subfloat[W$\to$A (PDA)]{
        \includegraphics[width=0.235\textwidth]{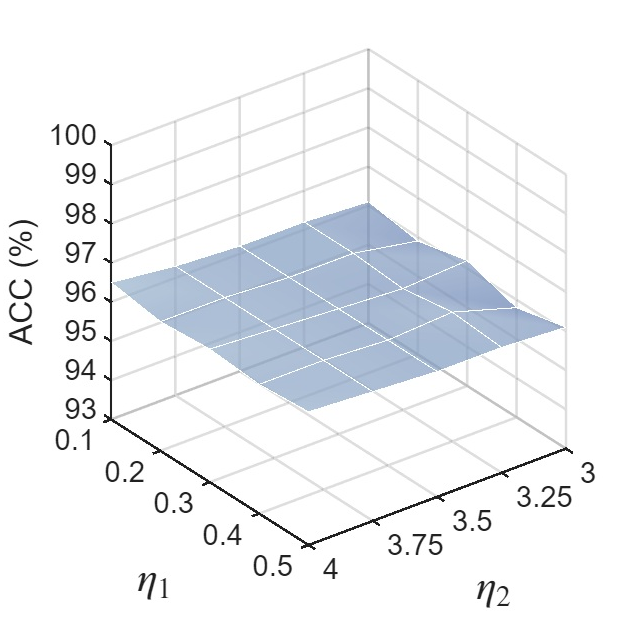}
        \label{fig:PDA_sensitivity_31_WtoA}}
    \subfloat[Cl$\to$Pr (PDA)]{
        \includegraphics[width=0.235\textwidth]{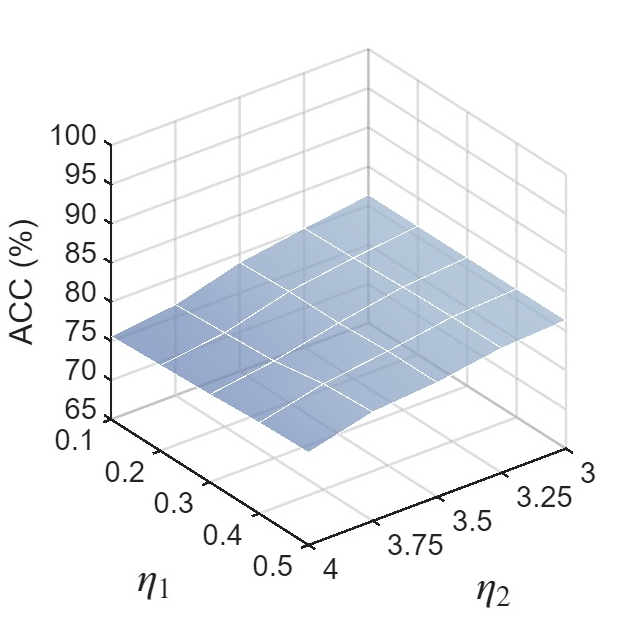}
        \label{fig:PDA_sensitivity_Home_WtoA}}
    \caption{Sensitivity analysis on hyper-parameters $\eta_1$ and $\eta_2$. (a)-(b) OSDA. (c)-(d) PDA.}
    \label{fig:Sensitivity}
\end{figure*}

\textbf{Effectiveness of the Private Class Identification.}
To verify the high efficiency of the proposed private class identification method—namely, that a majority of private class samples can be successfully identified—we conduct experiments on three benchmark datasets. To measure the efficiency, we calculate the average ratio of identified private class samples to all private class samples across all transfer tasks for each dataset. The experiment results are presented in Table~\ref{tab:private_identification}. We observe that the proposed private class identification approach identifies nearly all private class samples on the Image-CLEF dataset, achieving a 90.7$\%$ identification ratio in OSDA and 95.1$\%$ in PDA. Similarly, on the Office-31 dataset, it achieves a 94.5$\%$ identification ratio in OSDA and 99.4$\%$ in PDA. On the Office-Home dataset, although the identification performance is slightly reduced, the proposed approach still recognizes most private class samples, achieving a 79.5$\%$ identification ratio in OSDA and 92.2$\%$ in PDA. Overall, our method successfully identifies the majority of private class samples across all three datasets, with an average identification ratio of 88.2$\%$ in OSDA and 95.6$\%$ in PDA, thereby validating its high efficiency.

\begin{figure}[t] \centering
    \includegraphics[scale=0.65]{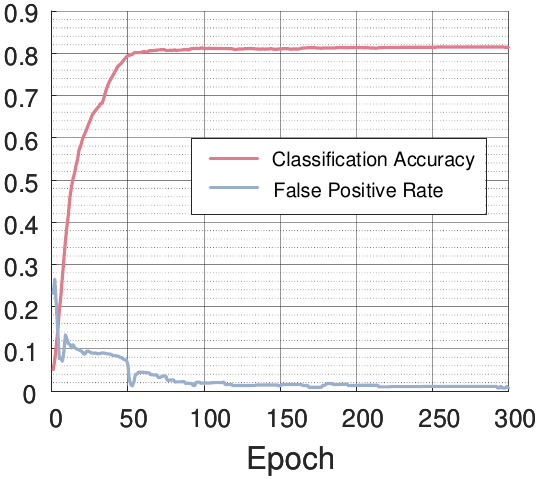}
    \vspace{-8pt}
    \caption{Model performance evolution across epochs on the Office-Home task Cl$\to$Pr. False positive rate denotes the proportion of shared-class samples that are mistakenly identified as private class samples.}
    \label{fig:false_positive_rate}
\end{figure}

To further assess the reliability of the private class identification, we also investigate the behavior of ReOT when there are no private classes. Specifically, on the Office-Home dataset, we remove all private classes and train the model using only the remaining 25 shared classes. The model is then evaluated on the target domain in terms of classification accuracy and false positive rate. The latter is obtained with our private class identification method and measures the proportion of shared class samples mistakenly identified as private. Fig.~\ref{fig:false_positive_rate} shows the result on task Cl$\to$Pr, ReOT achieves a high accuracy of about 82$\%$ while keeping the false positive rate consistently low throughout training. This behavior can be explained by the fact that, without private classes, shared samples naturally cluster around their class centers. Consequently, the transport mass within local regions becomes approximately uniform, resulting in consistently low identification scores and thus preventing false identification as private. These results indicate that even in the absence of private classes, ReOT rarely misidentifies shared class samples as private, validating the reliability of the proposed method.

\begin{table}[t]
\small
\renewcommand\arraystretch{1.0}
 \renewcommand\tabcolsep{0.45pc}
    \begin{center}
    \caption{The ratio ($\%$) of identified private class samples to all private class samples using the proposed private class identification method.}
        \label{tab:private_identification}
        \vspace{-1.0pc}
    \begin{tabular}{c|c|c|c|c}
        \toprule
        \textbf{Setting} & Image-CLEF & Office-31 & Office-Home & \textbf{Mean}\\
        \hline
        \rule{0pt}{14.5pt} 
        OSDA
        &90.7
        &94.5
        &79.5
        &88.2\\
        PDA
        &95.1
        &99.4
        &92.2
        &95.6\\

        \bottomrule
        \end{tabular}
    \end{center}
\end{table}

\textbf{Hyper-parameter Analysis.}
To investigate the sensitivity of the model w.r.t. hyper-parameters $\eta_1$ and $\eta_2$, we conduct experiments on Office-31 and Office-Home datasets. For the OSDA setting, both the values of $\eta_1$ and $\eta_2$ are selected from \{0.5, 0.75, 1, 1.25, 1.5\}. For the PDA setting, the values of $\eta_1$ and $\eta_2$ are searched from \{0.1, 0.2, 0.3, 0.4, 0.5\} and \{3, 3.25, 3.5, 3.75, 4\}, respectively. The 3-D grid visualizations for the results are presented in Figure~\ref{fig:Sensitivity}, where Office-31 task W$\to$A and Office-Home task Cl$\to$Pr are tested. It can be observed that ReOT is generally stable for different choices of parameters $\eta_1$ and $\eta_2$. Besides, the best performance is typically achieved with a larger value of $\eta_1$, further indicating the crucial role of invariant representation learning in both OSDA and PDA. In conclusion, the results above demonstrate that ReOT is generally robust to different settings of hyper-parameters.

\begin{figure*}[t] \centering
\hspace{2mm}
    \subfloat[Source-only]{
        \includegraphics[width=0.236\textwidth]{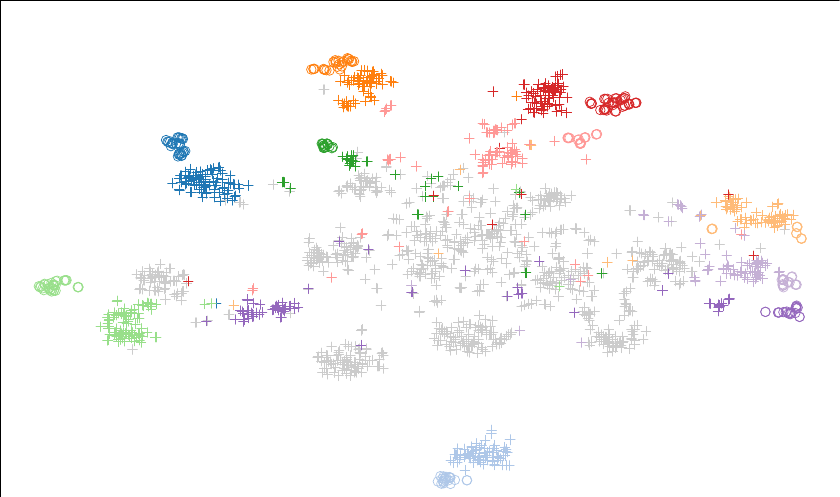}
        \label{fig:tSNE_original}}
    \subfloat[STA]{
        \includegraphics[width=0.236\textwidth]{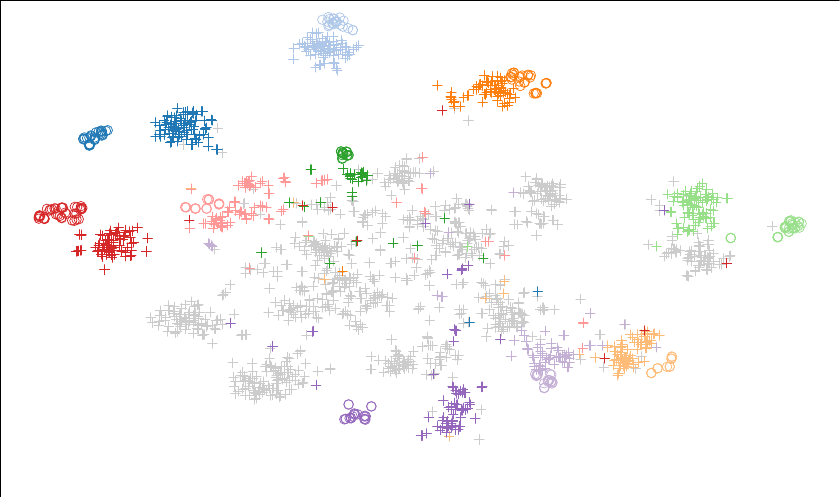}
        \label{fig:tSNE_STA}}
    \subfloat[ANNA]{
        \includegraphics[width=0.236\textwidth]{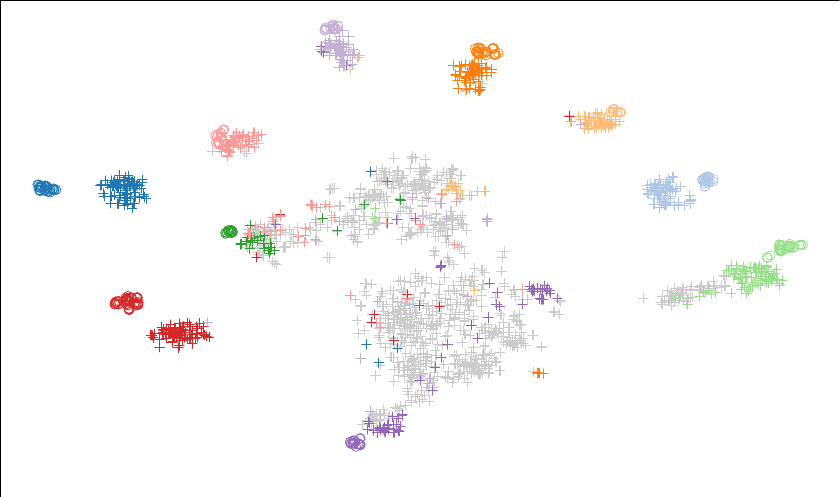}
        \label{fig:tSNE_ANNA}}
    \subfloat[ReOT]{
        \includegraphics[width=0.236\textwidth]{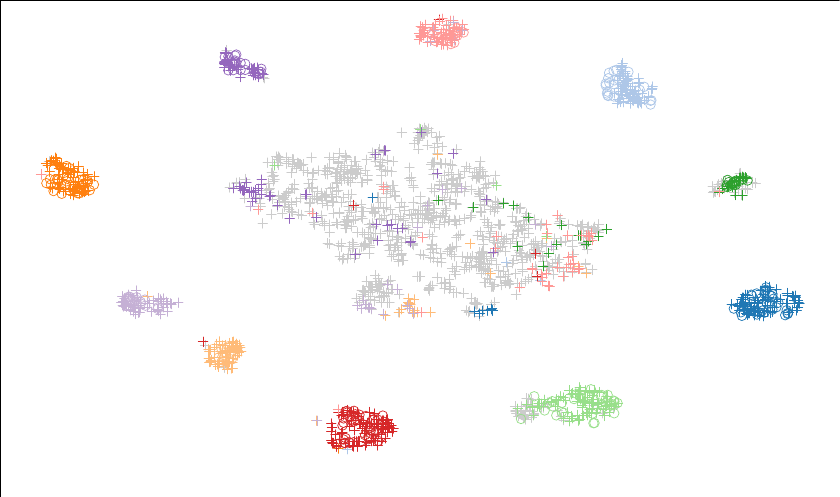}
        \label{fig:tSNE_ReOT}}

    \caption{t-SNE \citep{van2008t-sne} visualizations of the representations learned by different OSDA methods on the Office-31 task W$\to$A. In the diagrams, ``$\circ$'' means source domain, and ``+'' means target domain; different colors represent different classes, and gray represents the private class.}
    \label{fig:tSNE_visualization_OSDA}
\end{figure*}
\begin{figure*}[t] \centering
\hspace{2mm}

    \subfloat[Source-only]{
        \includegraphics[width=0.236\textwidth]{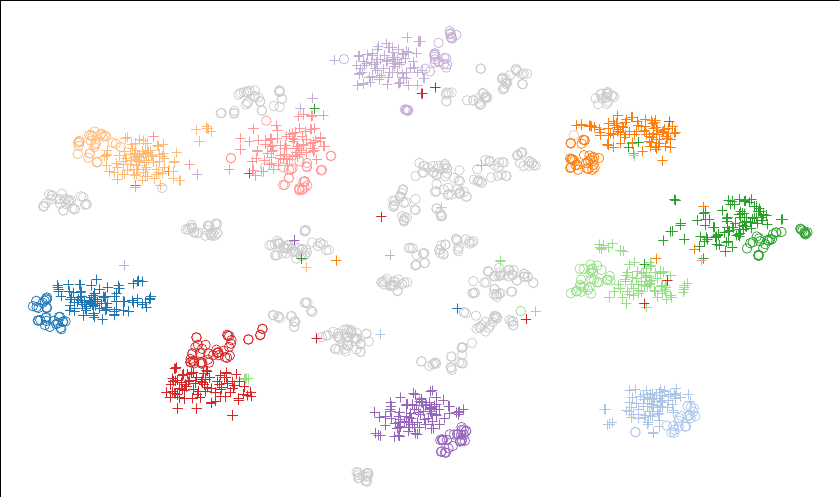}
        \label{fig:tSNE_PDA_Source}}
    \subfloat[AR]{
        \includegraphics[width=0.236\textwidth]{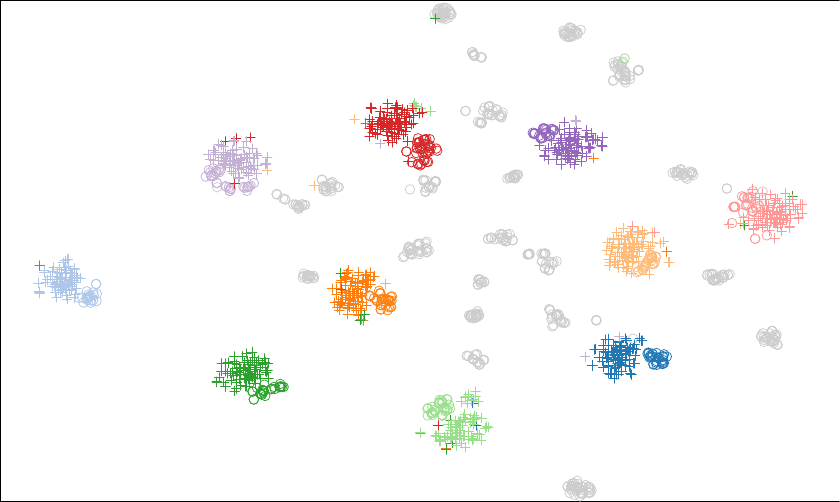}
        \label{fig:tSNE_PDA_AR}}
    \subfloat[SLM]{
        \includegraphics[width=0.236\textwidth]{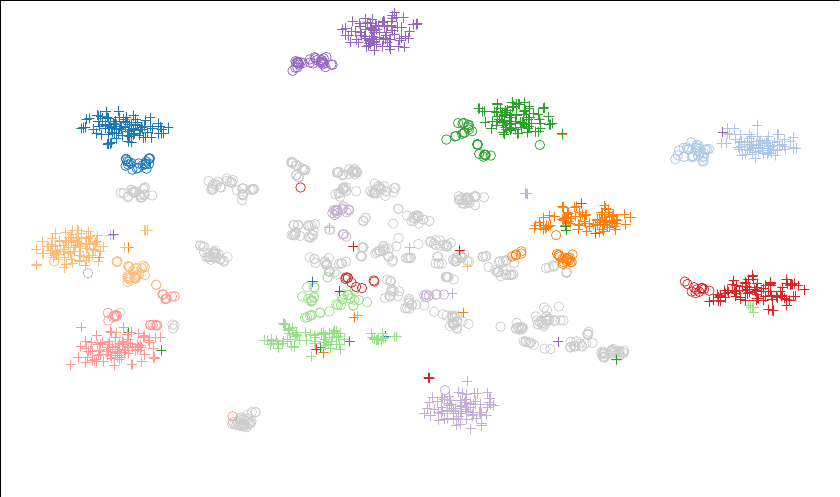}
        \label{fig:tSNE_PDA_SLM}}
    \subfloat[ReOT]{
        \includegraphics[width=0.236\textwidth]{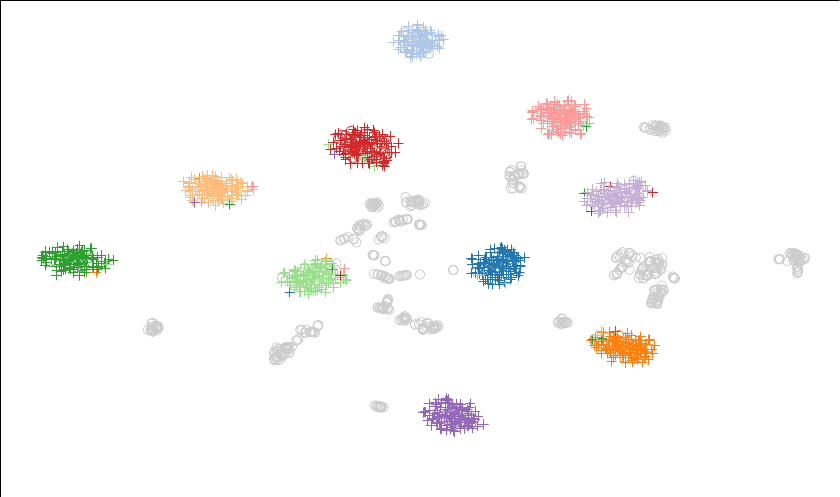}
        \label{fig:tSNE_PDA_Ours}}
    
    \caption{t-SNE \citep{van2008t-sne} visualizations of the representations learned by different PDA methods on the Office-31 task W$\to$A. In the diagrams, ``$\circ$'' means source domain, and ``+'' means target domain; different colors represent different classes, and gray represents the private class.}
    \label{fig:tSNE_visualization_PDA}
\end{figure*}

\begin{figure*}[!t] \centering
    \subfloat[OSDA]{
        \includegraphics[width=0.36\textwidth]{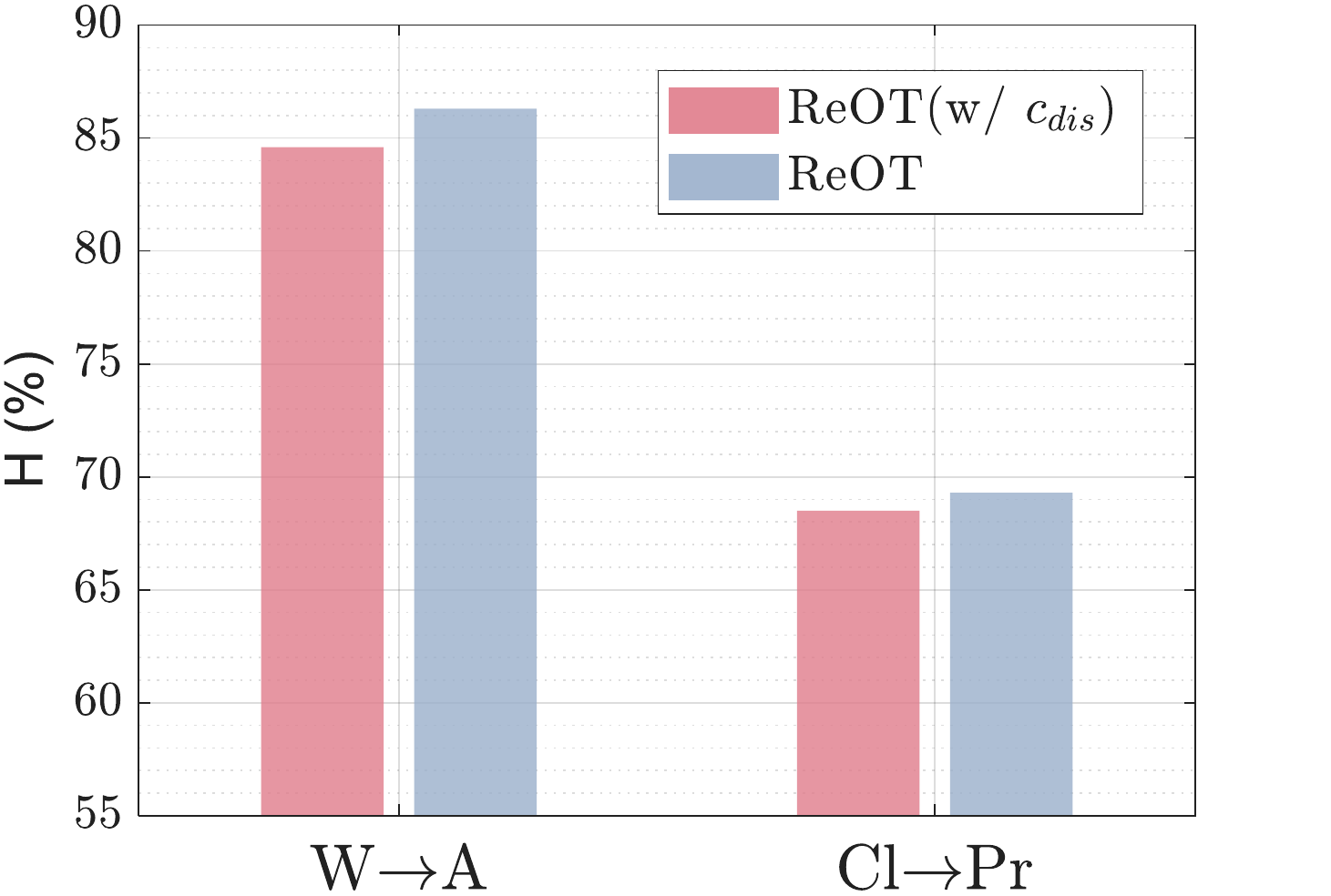}
        \label{fig:OSDA_cost_function}}\hspace{10mm}
    \subfloat[PDA]{
     \includegraphics[width=0.36\textwidth]{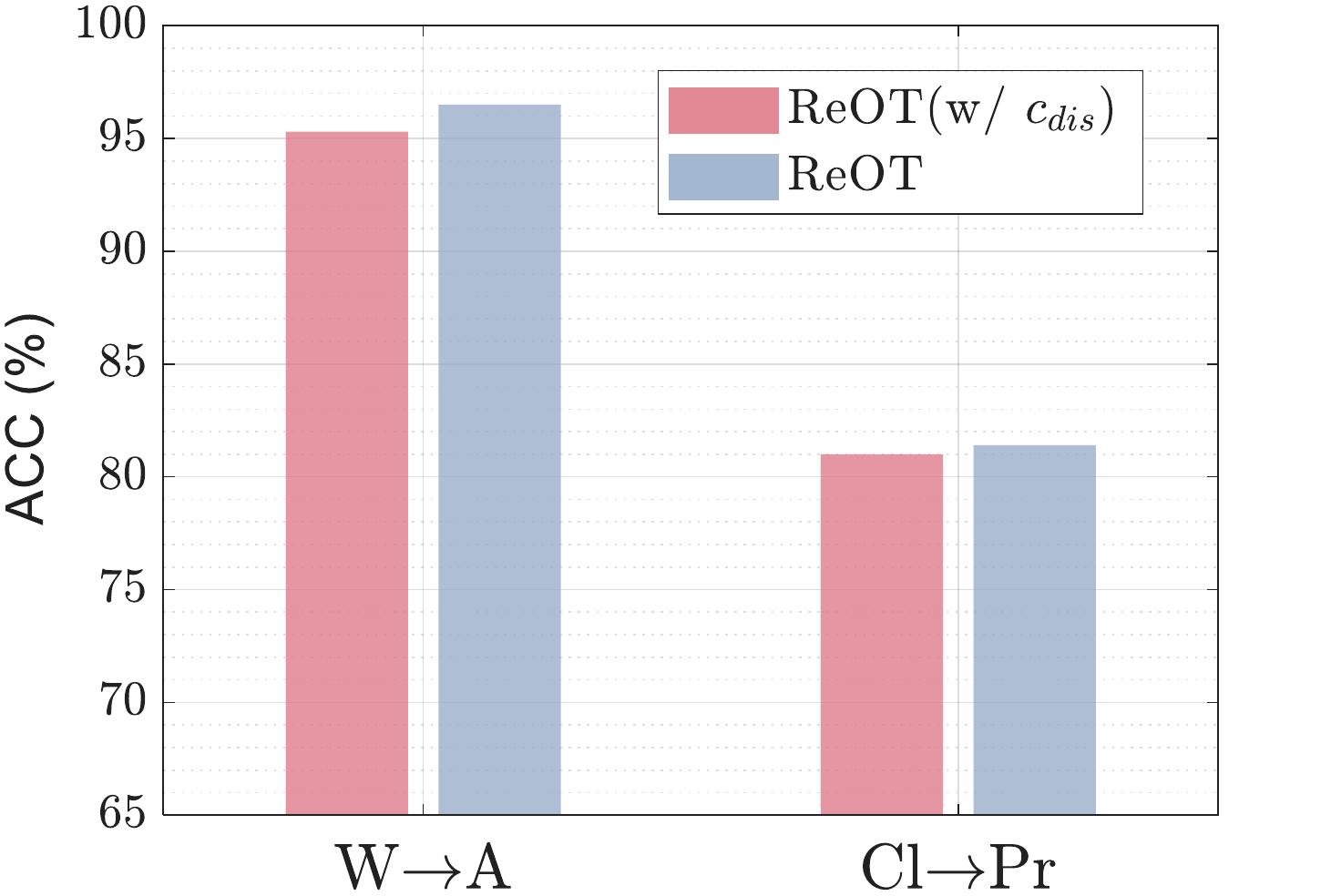}
        \label{fig:PDA_cost_function}}
    \caption{Impact of cost function selection in optimal transport on model performance. `` ReOT (w/ $c_{\mathrm{dis}}$) " indicates using a discriminator-based cost function, while `` ReOT " uses the default squared Euclidean cost. (a) OSDA setting. (b) PDA setting.}
    \label{fig:cost_function}
\end{figure*}

\textbf{Feature Visualization.}
As shown in Figures~\ref{fig:tSNE_visualization_OSDA} and~\ref{fig:tSNE_visualization_PDA}, we conduct t-SNE feature visulization and comparison for both OSDA and PDA. In the OSDA setting, as shown in Figure~\ref{fig:tSNE_original}, the target private class in gray is mixed with the shared classes before adaptation. Though STA and ANNA improve the discrimination of features, there is still a large overlap between private and shared classes, and the intra-class distance is not sufficiently small as shown in Figures~\ref{fig:tSNE_STA} and~\ref{fig:tSNE_ANNA}. Compared with these methods, ReOT separates the private and shared classes more completely and ensures well-aligned cluster structures as shown in Figure~\ref{fig:tSNE_ReOT}. Similarly, in the PDA setting, although AR (Figure~\ref{fig:tSNE_PDA_AR}) and SLM (Figure~\ref{fig:tSNE_PDA_SLM}) learn improved representation compared with Source-only (Figure~\ref{fig:tSNE_PDA_Source}), ReOT achieves better intra-class compactness and inter-class separability as shown in Figure~\ref{fig:tSNE_PDA_Ours}. These results demonstrate that ReOT ensures a better representation space, and further indicate the effectiveness of ReOT in mitigating the cross-domain class-conditional distribution discrepancy.   

\textbf{Impact of the Cost Function.}
To examine the robustness of ReOT with respect to the choice of the cost function in OT, we conduct additional ablation studies on the Office-31 and Office-Home datasets. Throughout our experiments, the default cost function $c$ is the squared Euclidean distance. Here, we compare it with a discriminator-based cost function $c_{dis}$. The discriminator $dis(\cdot)$ is a network that outputs a probability, indicating the likelihood that $\bm{z}$ belongs to the source domain. It is pre-trained using a binary cross-entropy on both source and target domain samples, and its parameters are frozen during the subsequent ReOT training. Intuitively, a source feature $\bm{z}$ and a target feature $\bm{z}'$ that are easily confused by the discriminator should have a low transport cost. Based on this, we define $c_{dis}$ as Based on this, we define $c_{dis}$ as:
\[
c_{dis}(\bm{z},\bm{z}') =|dis(\bm{z}) -0.5|+|dis(\bm{z}') -0.5|.
\]

Fig.~\ref{fig:cost_function} presents the results for the Office-31 task W$\to$A and the Office-Home task Cl$\to$Pr under both OSDA and PDA settings. The results show that ReOT with the default squared Euclidean cost consistently achieves superior performance across all settings. For example, on the W$\to$A task under the OSDA setting, ReOT outperforms ReOT(w/ $c_{dis}$) by approximately 1.7$\%$ on H. A similar trend is observed in the PDA setting, where ReOT also demonstrates better performance. Despite these performance differences, ReOT(w/ $c_{dis}$) still delivers competitive results. This indicates that the proposed method is robust and not sensitive to the selection of the cost function.

\section{Conclusion}
\label{sec5}
In this paper, we addressed the challenge of extreme label shift in domain adaptation by proposing a novel locality-aware private class identification method, which is achieved by defining a score function on masked optimal transport mass. This method relaxes the strong assumptions on which existing methods are based, and its effectiveness is demonstrated from theoretical perspective, highlighting its strong ability to distinguish between shared and private class samples. Building upon this foundation, we introduce the reliable OT-based (ReOT) method for practical applications. ReOT integrates risk minimization and employs masked transport to learn invariant representation with separated cluster structures. In addition, we provide a generalization bound of the target risk for the severe label shift scenario. This upper bound is shown to be tight and can be minimized by ReOT. Extensive experiments on benchmark datasets validated the effectiveness of ReOT, indicating its reliable and superior performance in extreme label shift scenarios.

While ReOT demonstrates strong performance in both OSDA and PDA, the current method focuses on the single-source single-target scenario and assumes access to labeled source data. A promising direction for future work is to extend ReOT to multi-source~\cite{zhao2020multi-source} and multi-target scenarios~\cite{isobe2021multi-target}, as well as to the source-free setting~\cite{liang2020source-free}. Potential strategies include generalizing the transport plan to a multi-marginal formulation~\cite{beier2023unbalanced-multi-marginal} for multiple domains, or using source class prototypes distilled from a pre-trained source model in source-free scenario~\cite{liang2020source-free}, while carefully addressing the associated computational and theoretical challenges.

\footnotesize
\bibliographystyle{IEEEtran}
\bibliography{ReOT_bib}

\end{document}